\documentclass[letterpaper]{article} 
\usepackage{aaai2027}
\usepackage[hyphens]{url}  
\usepackage{graphicx} 
\def\UrlFont{\rm}  
\usepackage{natbib}  
\usepackage{caption} 
\usepackage{algorithm}
\usepackage{algorithmic}
\usepackage{amsmath}
\usepackage{amssymb}

\usepackage{booktabs}
\usepackage{multirow}
\usepackage{graphicx}
\usepackage{newfloat}
\usepackage{listings}
\DeclareCaptionStyle{ruled}{labelfont=normalfont,labelsep=colon,strut=off} 
\floatstyle{ruled}
\newfloat{listing}{tb}{lst}{}
\floatname{listing}{Listing}

\usepackage{booktabs}

\title{DeCo-MIL: Debiased Counterfactual Reasoning for Long-Tailed Whole Slide Image Analysis}

\author{
    Xiaoxiao Li\textsuperscript{\rm 1}\equalcontrib,
    Xitong Ling\textsuperscript{\rm 1}\equalcontrib,
    Jiawen Li\textsuperscript{\rm 1}\equalcontrib,
    Weiming Chen\textsuperscript{\rm 1},
    Zhenyang Cai\textsuperscript{\rm 2},
    Xidong Wang\textsuperscript{\rm 2},
    Tian Guan\textsuperscript{\rm 1}\corresponding,
    Benyou Wang\textsuperscript{\rm 2}\corresponding,
    Yonghong He\textsuperscript{\rm 1}\corresponding
}
\affiliations{
    \textsuperscript{\rm 1}Tsinghua University\\
    \textsuperscript{\rm 2}The Chinese University of Hong Kong, Shenzhen\\
}

\begin{document}

\maketitle

\begin{abstract}
Multiple instance learning (MIL) is widely used for weakly supervised whole slide image (WSI) analysis. However, under long-tailed distributions, MIL-based WSI analysis faces a nested dual long-tail: an inter-slide class long tail and an intra-slide long tail of instance-level discriminative evidence. The two long tails are coupled: tail classes have few training slides, while their limited diagnostic evidence is concentrated in a few patches and obscured by abundant within-bag redundancy. This coupling biases models toward head classes and degrades rare-class recognition. To address this, we propose DeCo-MIL for long-tailed WSI analysis, which jointly alleviates the nested dual long-tail through frequency-debiased counterfactual reasoning. For the inner long tail, DeCo-MIL clusters patches into tissue-morphology anchors, replaces each anchor with its matched normal prototype to perform a counterfactual intervention, and estimates its counterfactual contribution to the ground-truth class using class-frequency-corrected predictions. These contributions guide redundancy masking to preserve scarce discriminative instances. For the outer long tail, DeCo-MIL constructs anchor-stratified pseudo-bags from redundancy-reduced bags and combines tail-aware oversampling with consistency regularization, increasing effective supervision for tail classes while preserving tissue-morphology composition. Extensive experiments on three long-tailed WSI benchmarks demonstrate that DeCo-MIL achieves state-of-the-art performance in both tail-class recognition and overall classification.

\end{abstract}



\section{Introduction}
\label{sec:intro}


Advances in deep learning and computing power have driven the adoption of computational pathology in diagnosis. However, a whole slide image (WSI) is a gigapixel pathology scan that cannot be processed at full resolution, while pixel- or region-level annotation is prohibitively expensive. Multiple instance learning (MIL) has thus become the dominant paradigm for WSI analysis using only slide-level annotations. In MIL, a WSI is tessellated into thousands of patches after non-tissue background removal; the patches are then encoded by a pretrained encoder and aggregated into a bag-level representation for classification.

\begin{figure}[t]
\centering
\includegraphics[width=1.0\columnwidth]{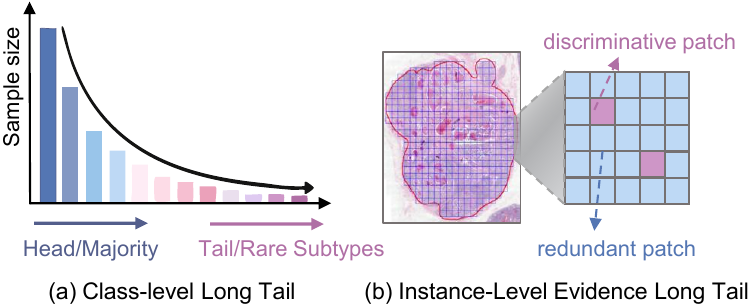} 
\caption{The nested dual long-tail: 
(a) the inter-slide class long tail, (b) the intra-slide long tail of instance-level discriminative evidence.}
\label{fig1}
\end{figure}

Research on MIL for WSIs advances along two complementary directions---stronger
patch representations and more expressive aggregation---steadily improving
overall accuracy. These advances, however, implicitly assume a balanced label
distribution. In real clinical data, diagnostic categories are heavily
long-tailed: common subtypes dominate while rare yet critical categories have
few samples. Under such skew, MIL models overfit head classes and underfit tail
classes, weakening tail recognition and overall reliability.

We argue that the difficulty of long-tailed WSI classification does not arise from a single source of imbalance, but from a \textbf{nested dual long-tail}. At the outer level, the numbers of slides across diagnostic categories follow a long-tailed distribution. At the inner level, instance-level discriminative evidence within each slide is also long-tailed: only a small subset of instances in a bag carries diagnostic information, while most correspond to common tissue patterns or redundant, non-discriminative content. 
The two levels are therefore not independent: tail classes already contain fewer training slides, while their limited discriminative evidence is further obscured by abundant redundant instances within each bag. This coupling makes slide-level class rebalancing alone insufficient to fully recover tail-class performance.

Existing long-tailed WSI methods, including MDE-MIL~\cite{mdemil} and SC-MIL~\cite{scmil}, improve tail recognition through multimodal ensemble learning or supervised contrastive learning. However, how to explicitly identify and preserve sparse diagnostic morphologies within tail-class bags remains underexplored.
A natural way to expose this evidence is to score each region by counterfactual attribution—replacing a region and observing how the prediction changes; however, under a long tail the model's posterior is distorted by class frequency, so reading it directly keeps tail evidence hidden, and the attribution must be performed after debiasing.

We therefore propose \textbf{DeCo-MIL}, a WSI classification method that jointly addresses the nested dual long-tail problem through debiased counterfactual reasoning. At the inner level, we perform counterfactual interventions by replacing each tissue anchor with a matched normal prototype and use frequency-debiased counterfactual contributions to guide redundancy masking, thereby uncovering buried tail-class evidence. At the outer level, we construct anchor-stratified pseudo-bags from the redundancy-reduced subset and oversample tail-class slides, further amplifying at the slide level the tail evidence recovered in the preceding step. 

Our contributions are summarized as follows:
\begin{itemize}

\item We characterize long-tailed WSI classification as a nested dual long-tail and empirically reveal the coupling between its class-level and instance-level evidence long tails.

\item We introduce debiased counterfactual-guided redundancy masking to identify and preserve scarce intra-slide discriminative evidence.

\item We develop tail-aware stratified pseudo-bag augmentation to alleviate the inter-slide class long tail.

\item Experiments on three long-tailed WSI benchmarks demonstrate that our method outperforms the baseline methods in both tail-class and overall performance.

\end{itemize}


\section{Related Work}
\label{sec:related}

\subsection{Multiple Instance Learning for WSIs}

MIL studies for WSI classification have largely focused on improving bag
aggregation. Instance aggregation has evolved from prior-based non-parametric
pooling to learnable attention mechanisms that
assign adaptive weights to instances, such as
AB-MIL~\cite{abmil} and CLAM~\cite{clam}. To capture long-range dependencies and model global context,
self-attention- and Transformer-based aggregators have also been widely
explored, including TransMIL~\cite{transmil} and FR-MIL~\cite{frmil}.
Complementary directions include graph-based methods for modeling relational
dependencies among patches, such as H$^2$-MIL~\cite{h2mil} and WiKG~\cite{wikg}, and clustering-based
methods, such as MiCo~\cite{micomil}, PANTHER~\cite{panther} and PAMoE~\cite{pamoe}.

Other studies suppress uninformative regions through redundancy removal or
hard-instance mining, including masked hard-instance mining
(MHIM-MIL~\cite{mhim}), attention-driven selection
(AMD-MIL~\cite{amdmil}), and saliency-guided masking
(SAM-MIL~\cite{sammil}).
However, these methods generally use class-agnostic selection criteria, identifying broadly salient regions rather than class-specific evidence and potentially overlooking scarce tail-class evidence.

\subsection{Long-Tailed Learning}

In natural images, mainstream approaches to the long-tailed problem span
resampling~\cite{buda2018systematic},
margin-based loss rebalancing~\cite{cao2019ldam}, 
data augmentation~\cite{chou2020remix},
logit adjustment~\cite{ren2020balancedsoftmax,menon2021logit},
transfer learning~\cite{wang2017modeltail},
decoupled training~\cite{kang2020decoupling},
and ensemble learning~\cite{wang2021ride}.
Recent work brings these into WSIs: MDE-MIL~\cite{mdemil} addresses long-tailed
pathology with multimodal distillation-driven ensemble learning, and
SC-MIL~\cite{scmil} handles imbalanced pathology with supervised contrastive
learning. 
These methods improve representations and decision boundaries under imbalanced slide distributions, while explicit morphology-level evidence estimation and redundancy control within tail-class bags remain underexplored.

\subsection{Counterfactual Reasoning}

Counterfactual reasoning has been widely used to analyze and improve deep models. For explanation and attribution, occlusion~\cite{zeiler2014} and meaningful-perturbation~\cite{fong2017} methods measure the importance of an input region by masking or replacing it and observing the output change, and SHAP~\cite{shap} unifies such feature attributions through Shapley values. For debiasing, CAL~\cite{cal} suppresses spurious focus by contrasting factual and counterfactual attention, whereas De-confound-TDE~\cite{deconfoundtde} treats the head-class momentum as a confounder and removes its direct effect at inference to improve tail classes in long-tailed recognition. In computational pathology, CIMIL~\cite{cimil} uses counterfactual-inference-based sub-bag evaluation and hierarchical instance searching for binary tumor detection, while IBMIL~\cite{ibmil} performs backdoor adjustment at the bag-representation level to deconfound contextual priors. However, none of these works is specifically tailored to long-tailed classification.

\begin{figure*}[t]
\centering
\includegraphics[width=1.0\textwidth]{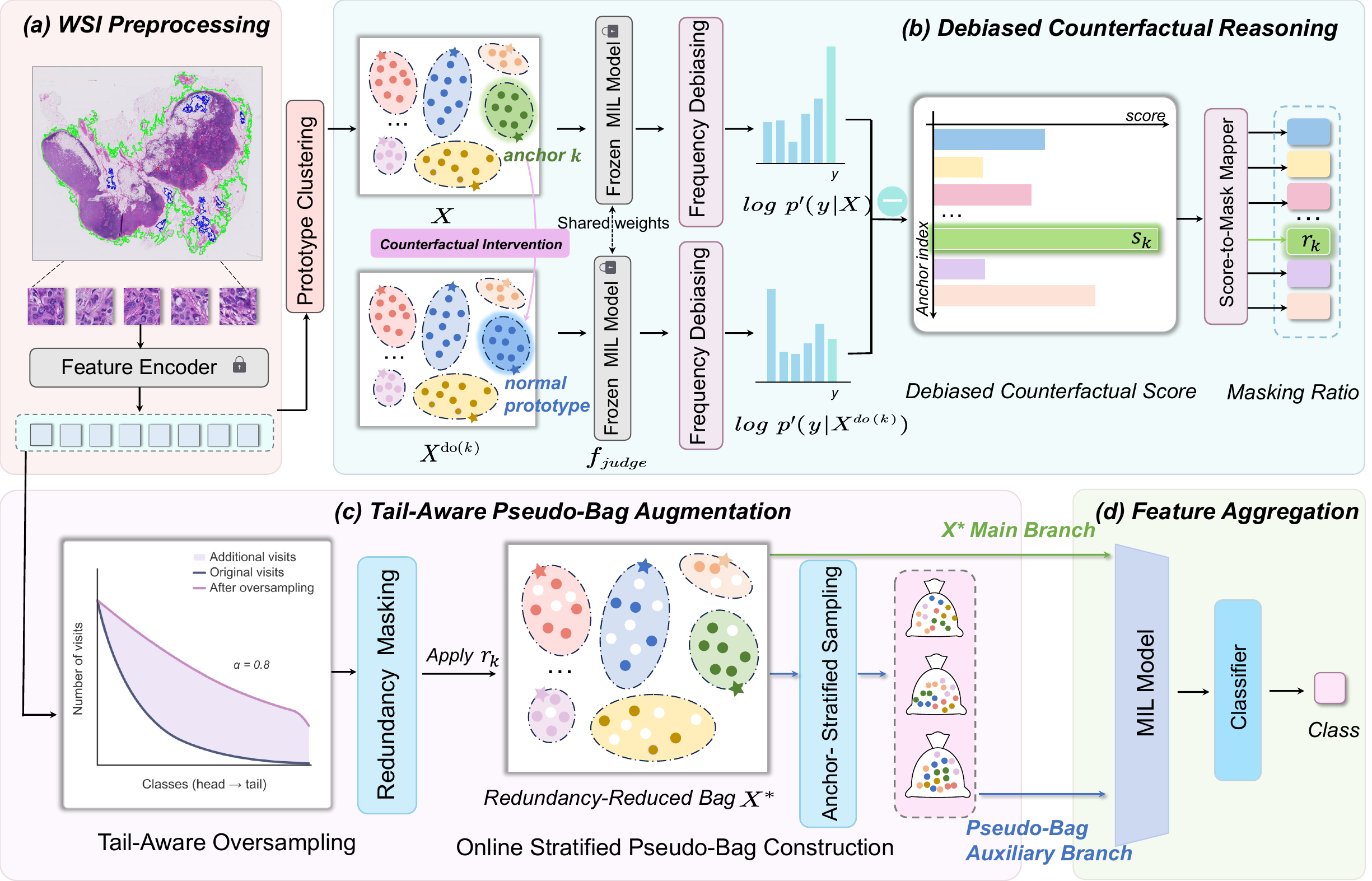} 
\caption{Overview of the DeCo-MIL training framework. (a) WSI preprocessing and prototype clustering. (b) Debiased counterfactual scoring maps anchor contributions to masking ratios. (c) Tail-class oversampling and anchor-stratified sampling construct pseudo-bags. (d) The main and pseudo-bag auxiliary branches jointly train the final MIL model.}
\label{fig2}
\end{figure*}


\section{Method}
\label{sec:method}

\subsection{Problem Formulation}
\label{sec:problem}

After background removal, a whole-slide image (WSI) is divided into a large number of patches.
Multiple instance learning (MIL) represents each slide as a bag of instances and uses only slide-level supervision. Let $\mathcal{D}=\{(X_i,y_i)\}_{i=1}^{M}$ denote the training set, where $X_i$ contains $N_i$ patches and $y_i\in\{1,\ldots,C\}$ is the slide-level label. A frozen encoder $\phi$ maps the $n$-th patch to an $\ell_2$-normalized feature:

\begin{equation}
x_{i,n}
=
\frac{\phi(\mathrm{patch}_{i,n})}
{\lVert\phi(\mathrm{patch}_{i,n})\rVert_2}
\in\mathbb{R}^{d}.
\label{eq:patch-feature}
\end{equation}

The resulting bag is $X_i=\{x_{i,n}\}_{n=1}^{N_i}$. An MIL aggregator $g$ summarizes the instances into a slide representation, from which a classifier $h$ predicts the class probabilities:

\begin{equation}
p_i
=
\operatorname{softmax}
\left(
h\left(g(X_i)\right)
\right).
\label{eq:mil-prediction}
\end{equation}


\paragraph{Inter-slide Class Long Tail.}
Let $n_c$ be the number of training slides in class $c$, with $n_1\geq n_2\geq\cdots\geq n_C$. We measure the skew of the class distribution by the imbalance ratio

\begin{equation}
\mathrm{IR}
=
\frac{\max_c n_c}{\min_c n_c}.
\label{eq:imbalance-ratio}
\end{equation}

When $\mathrm{IR}$ is large, slide-level supervision is dominated by head classes, biasing the model against tail classes.

\paragraph{Intra-slide Long Tail of Instance-level Discriminative Evidence.}


Within each instance bag, the discriminative contributions of individual instances follow a highly skewed, long-tailed distribution: 
only a small fraction of patches carry the dominant diagnostic evidence, whereas the majority correspond to normal tissue or non-discriminative redundancy. Since instance-level annotations are unavailable, this evidence distribution cannot be directly observed. The subsequent Visualization and Analysis section provides empirical support for the intra-slide long tail of instance-level discriminative evidence and its coupling with the class-level long tail through an analysis of debiased counterfactual contributions.


\subsection{Method Overview}
\label{sec:overview}

As illustrated in Fig.~\ref{fig2}, DeCo-MIL follows three
stages. In Stage~0, a standard MIL model is trained on the
original bags and frozen as $f_{\mathrm{judge}}$. In Stage~1,
the frozen judge estimates debiased counterfactual
contributions for shared morphology anchors and maps them
to anchor-specific masking ratios. In Stage~2, these masking
ratios are used to construct redundancy-reduced bags and
anchor-stratified pseudo-bags for training
$f_{\mathrm{final}}$ with tail-aware oversampling and
consistency regularization. Only $f_{\mathrm{final}}$ is used for
inference.

\subsection{Debiased Counterfactual-Guided Redundancy Masking}
\label{sec:counterfactual}

DeCo-MIL intervenes on groups of morphologically related instances rather than individual patches. We apply global K-means clustering to all training patch features and obtain $K=64$ fixed morphology prototypes $\{H_k\}_{k=1}^{K}$. Each patch is assigned to its nearest anchor by cosine similarity: 

\begin{equation}
z_n
=
\operatorname*{arg\,max}_{k}
\cos(x_n,H_k).
\label{eq:anchor-assignment}
\end{equation}

Because the prototypes are shared across the training set, they provide a common feature-space reference across slides. We additionally cluster patches from normal slides and select, for each $H_k$, the most similar normal prototype $\hat{q}_k$.

For anchor $k$ in slide $X$, we perform a model-level counterfactual intervention $\mathrm{do}(x_n=\hat{q}_k)$ for all instances assigned to that anchor while leaving the remaining instances unchanged. Let $X^{\mathrm{do}(k)}=\{\tilde{x}_{n}^{(k)}\}_{n=1}^{N}$ denote the intervened bag, where

\begin{equation}
\tilde{x}_{n}^{(k)}
=
\begin{cases}
\hat{q}_k, & z_n=k,\\
x_n, & z_n\neq k.
\end{cases}
\label{eq:counterfactual-bag}
\end{equation}

This intervention asks how the frozen model's diagnosis would change if the $k$-th morphology were replaced by its matched normal counterpart. Compared with zeroing or random noise, normal-prototype replacement helps reduce intervention-induced feature shift.

We next evaluate the prediction change using the frozen judge model. Since class frequency can bias its posterior toward head classes, we calibrate the judge logits using the training prior $\pi_c=n_c/\sum_{c'}n_{c'}$, following prior-aware logit
adjustment~\cite{menon2021logit}:

\begin{equation}
p'(c \mid X) =
\operatorname{softmax}\!\left( f_{\mathrm{judge}}(X) - \tau \log \pi \right)_{c}.
\label{eq:frequency-calibration}
\end{equation}

Here, $\tau$ controls the calibration strength, and is set to
$1.0$ for all datasets. The debiased counterfactual contribution of anchor $k$ to  class
$c$ is the difference in debiased log-probability before and after the
intervention:

\begin{equation}
{E'}_{k}^{\,c}
=
\log p'(c \mid X)
-
\log p'\!\left(c \mid X^{\mathrm{do}(k)}\right).
\label{eq:counterfactual-contribution}
\end{equation}

For the ground-truth class $y$, a large ${E'}_{k}^{\,y}$ indicates that replacing anchor $k$ substantially decreases the probability of the ground-truth class, suggesting that the anchor carries discriminative evidence. A contribution close to zero instead indicates likely redundancy.

To penalize non-specific anchors that also support competing classes, we define
$\tilde{y}=\arg\max_{c\neq y}p'(c\mid X)$ as the highest-probability
non-ground-truth class under the frequency-calibrated prediction of the
original slide.
We define the anchor score as:

\begin{equation}
s_k = {E'}_{k}^{\,y} - \lambda\, {E'}_{k}^{\,\tilde{y}}.
\label{eq:anchor-score}
\end{equation}


Equivalently, $s_k=\lambda(E_k^{\prime y}-E_k^{\prime\tilde{y}})
+(1-\lambda)E_k^{\prime y}$ balances class-specific contrast with
frequency-calibrated ground-truth contribution. Based on sensitivity analysis, we set $\lambda=0.6$ for all datasets,
thereby retaining the frequency-calibrated normalization term in $s_k$.

We apply slide-wise min-max normalization to the anchor scores, denoted by $\mathrm{Norm}(s_k)$, with $\epsilon=10^{-8}$ added to the denominator for
numerical stability, and map them to masking ratios:

\begin{equation}
r_k
=
r_{\max}
\left[
1-\mathrm{Norm}(s_k)
\right].
\label{eq:mask-ratio}
\end{equation}

Here, $r_{\max}<1$ is the maximum masking ratio. High-contribution anchors receive smaller masking ratios and are preferentially retained, whereas low-contribution anchors are more aggressively removed. We set $r_k=0$ for normal slides. All masking ratios are computed offline before final model training; instance-level random removal is performed online as described next.

\subsection{Tail-Aware Pseudo-Bag Augmentation}
\label{sec:pseudo-bag}

Although the preceding redundancy-masking stage recovers discriminative evidence within each bag, tail classes still suffer from limited slide-level samples. We therefore construct stratified pseudo-bags from redundancy-reduced bags and combine them with class-aware oversampling.

For slide $X$, let $G_k$ be the instance set assigned to anchor $k$, and let
$\mathcal K=\{k:G_k\neq\varnothing\}$
denote the set of its non-empty anchor indices. Based on $r_k$, we randomly retain instances from each anchor to obtain the redundancy-reduced bag:


\begin{equation}
\begin{aligned}
X^{*}
=
\bigcup_{k\in\mathcal{K}}
\operatorname{RandSubset}
\Bigl(
G_k,\,
\bigl\lfloor(1-r_k)|G_k|\bigr\rceil
\Bigr).
\end{aligned}
\label{eq:reduced-bag}
\end{equation}

Here, $\operatorname{RandSubset}(G,m)$ uniformly samples $m$ instances  without replacement, and
$\lfloor\cdot\rceil$ denotes nearest-integer rounding. With fixed masking
ratios, retained instances are resampled at each visit,
providing instance-level augmentation.


We then construct pseudo-bags by stratified sampling over
the retained anchors. Let $G_k^{*}$ be the instances of anchor
$k$ retained in $X^{*}$, and let
$\mathcal K^{*}=\{k\in\mathcal K:G_k^{*}\neq\emptyset\}$.
The $m$-th pseudo-bag is


\begin{equation}
B^{(m)}
=
\bigcup_{\!k\in\mathcal K^\star}
\operatorname{RandSubset}\!\left(
G_k^\star,\max\{1,\lfloor\rho|G_k^\star|\rceil\}
\right),
\label{eq:pseudo-bag}
\end{equation}

where the sampling ratio is fixed at $\rho=0.5$. The lower bound of one prevents any retained anchor from being omitted, reducing the risk that a discriminative anchor is absent from a pseudo-bag. At every visit, we independently generate $M_{\mathrm{pb}}=3$ pseudo-bags inheriting the source-slide
label $y$.

Class-level rebalancing is implemented through tail-aware oversampling. Let $n_{\max}$ be the size of the largest class. A slide from class $c$ is sampled $a_c$ times per epoch:

\begin{equation}
a_c
=
\operatorname{clip}
\left(
\operatorname{round}
\left[
\left(\frac{n_{\max}}{n_c}\right)^{\alpha}
\right],
1,\mathrm{cap}
\right),
\label{eq:tail-oversampling}
\end{equation}

 where $\alpha$ controls the oversampling strength and $\mathrm{cap}$ limits the maximum number of repetitions; we
set $\mathrm{cap}=8$ across datasets. Tail-class slides are thus sampled more frequently, and each sampling generates different redundancy-reduced bags and pseudo-bags. The resulting $X^*$ and pseudo-bags are used by the main and auxiliary branches, respectively.

\begin{table*}[t]
  \centering
  \setlength{\tabcolsep}{2.8pt}
  \renewcommand{\arraystretch}{1.12}
  \resizebox{\textwidth}{!}{%
  \begin{tabular}{l*{12}{c}}
    \toprule
    \multirow{2}{*}{Method}
    & \multicolumn{4}{c}{Camelyon+-LT (IR=30)}
    & \multicolumn{4}{c}{PANDA-LT (IR=50)}
    & \multicolumn{4}{c}{TissueNet-LT (IR=10)} \\
    \cmidrule(lr){2-5}
    \cmidrule(lr){6-9}
    \cmidrule(lr){10-13}
    & ACC(\%) & AUC(\%) & F1(\%) & Tail-F1(\%)
    & ACC(\%) & AUC(\%) & F1(\%) & Tail-F1(\%)
    & ACC(\%) & AUC(\%) & F1(\%) & Tail-F1(\%) \\
    \midrule

    AB-MIL
    & $61.56_{\scriptscriptstyle 2.54}$
    & $87.76_{\scriptscriptstyle 1.69}$
    & $52.46_{\scriptscriptstyle 2.71}$
    & $0.00_{\scriptscriptstyle 0.00}$
    & $51.28_{\scriptscriptstyle 0.83}$
    & $86.74_{\scriptscriptstyle 0.36}$
    & $45.90_{\scriptscriptstyle 1.30}$
    & $19.85_{\scriptscriptstyle 3.38}$
    & $56.67_{\scriptscriptstyle 1.05}$
    & $83.06_{\scriptscriptstyle 0.96}$
    & $49.64_{\scriptscriptstyle 0.81}$
    & $0.00_{\scriptscriptstyle 0.00}$ \\

    CLAM-SB
    & $63.75_{\scriptscriptstyle 1.53}$
    & $89.43_{\scriptscriptstyle 0.59}$
    & $54.54_{\scriptscriptstyle 1.61}$
    & $0.00_{\scriptscriptstyle 0.00}$
    & $49.48_{\scriptscriptstyle 0.50}$
    & $86.80_{\scriptscriptstyle 0.32}$
    & $41.74_{\scriptscriptstyle 0.84}$
    & $6.76_{\scriptscriptstyle 3.22}$
    & $55.50_{\scriptscriptstyle 0.41}$
    & $81.96_{\scriptscriptstyle 0.43}$
    & $48.42_{\scriptscriptstyle 0.30}$
    & $0.00_{\scriptscriptstyle 0.00}$ \\

    CLAM-MB
    & $61.88_{\scriptscriptstyle 0.77}$
    & $80.46_{\scriptscriptstyle 0.36}$
    & $52.84_{\scriptscriptstyle 0.82}$
    & $0.00_{\scriptscriptstyle 0.00}$
    & $51.40_{\scriptscriptstyle 0.84}$
    & $86.93_{\scriptscriptstyle 0.28}$
    & $45.70_{\scriptscriptstyle 1.68}$
    & $18.40_{\scriptscriptstyle 5.21}$
    & $59.50_{\scriptscriptstyle 0.85}$
    & $83.82_{\scriptscriptstyle 0.57}$
    & $52.75_{\scriptscriptstyle 1.47}$
    & $3.67_{\scriptscriptstyle 4.87}$ \\

    DS-MIL
    & $64.69_{\scriptscriptstyle 3.64}$
    & $88.65_{\scriptscriptstyle 2.31}$
    & $55.67_{\scriptscriptstyle 3.55}$
    & $0.00_{\scriptscriptstyle 0.00}$
    & $57.40_{\scriptscriptstyle 1.23}$
    & $88.32_{\scriptscriptstyle 0.35}$
    & $55.18_{\scriptscriptstyle 1.50}$
    & $41.68_{\scriptscriptstyle 6.15}$
    & $62.83_{\scriptscriptstyle 2.51}$
    & $86.16_{\scriptscriptstyle 1.86}$
    & $57.47_{\scriptscriptstyle 1.70}$
    & $15.48_{\scriptscriptstyle 4.37}$ \\

    TransMIL
    & $63.12_{\scriptscriptstyle 3.06}$
    & $83.82_{\scriptscriptstyle 2.48}$
    & $56.59_{\scriptscriptstyle 4.30}$
    & $14.37_{\scriptscriptstyle 9.74}$
    & $53.35_{\scriptscriptstyle 1.49}$
    & $85.46_{\scriptscriptstyle 1.10}$
    & $51.08_{\scriptscriptstyle 2.31}$
    & $40.61_{\scriptscriptstyle 8.21}$
    & $63.33_{\scriptscriptstyle 2.64}$
    & $86.36_{\scriptscriptstyle 1.61}$
    & $60.53_{\scriptscriptstyle 3.29}$
    & $26.96_{\scriptscriptstyle 10.54}$ \\

    ILRA-MIL
    & $64.69_{\scriptscriptstyle 3.22}$
    & $88.73_{\scriptscriptstyle 1.27}$
    & $56.77_{\scriptscriptstyle 4.22}$
    & $6.93_{\scriptscriptstyle 5.66}$
    & $58.02_{\scriptscriptstyle 0.85}$
    & $88.47_{\scriptscriptstyle 0.32}$
    & $56.60_{\scriptscriptstyle 0.69}$
    & $45.67_{\scriptscriptstyle 2.82}$
    & $62.83_{\scriptscriptstyle 2.27}$
    & $84.66_{\scriptscriptstyle 0.98}$
    & $60.96_{\scriptscriptstyle 2.87}$
    & $35.42_{\scriptscriptstyle 7.23}$ \\

    MHIM-MIL
    & $62.19_{\scriptscriptstyle 1.53}$
    & $88.17_{\scriptscriptstyle 1.09}$
    & $53.15_{\scriptscriptstyle 1.96}$
    & $0.00_{\scriptscriptstyle 0.00}$
    & $56.85_{\scriptscriptstyle 0.77}$
    & $87.62_{\scriptscriptstyle 0.32}$
    & $55.38_{\scriptscriptstyle 1.38}$
    & $45.04_{\scriptscriptstyle 3.31}$
    & $63.50_{\scriptscriptstyle 2.91}$
    & $84.99_{\scriptscriptstyle 1.45}$
    & $59.35_{\scriptscriptstyle 4.34}$
    & $20.67_{\scriptscriptstyle 13.81}$ \\

    MiCo-MIL
    & $60.00_{\scriptscriptstyle 1.59}$
    & $87.27_{\scriptscriptstyle 1.38}$
    & $51.21_{\scriptscriptstyle 1.44}$
    & $4.71_{\scriptscriptstyle 5.76}$
    & $54.82_{\scriptscriptstyle 2.17}$
    & $87.04_{\scriptscriptstyle 0.69}$
    & $52.46_{\scriptscriptstyle 2.51}$
    & $37.86_{\scriptscriptstyle 5.81}$
    & $66.83_{\scriptscriptstyle 3.63}$
    & $87.72_{\scriptscriptstyle 0.65}$
    & $65.09_{\scriptscriptstyle 4.36}$
    & $41.01_{\scriptscriptstyle 8.38}$ \\

    AMD-MIL
    & $61.56_{\scriptscriptstyle 2.72}$
    & $84.74_{\scriptscriptstyle 3.29}$
    & $54.16_{\scriptscriptstyle 2.50}$
    & $14.60_{\scriptscriptstyle 12.15}$
    & $57.02_{\scriptscriptstyle 1.09}$
    & $87.74_{\scriptscriptstyle 0.42}$
    & $55.18_{\scriptscriptstyle 1.53}$
    & $43.86_{\scriptscriptstyle 3.25}$
    & $63.67_{\scriptscriptstyle 1.35}$
    & $85.50_{\scriptscriptstyle 0.60}$
    & $62.53_{\scriptscriptstyle 1.11}$
    & $36.81_{\scriptscriptstyle 3.44}$ \\

    WiKG-MIL
    & $63.12_{\scriptscriptstyle 3.37}$
    & $89.38_{\scriptscriptstyle 2.56}$
    & $57.00_{\scriptscriptstyle 4.09}$
    & $15.82_{\scriptscriptstyle 5.24}$
    & $59.27_{\scriptscriptstyle 1.84}$
    & $88.55_{\scriptscriptstyle 0.73}$
    & $57.81_{\scriptscriptstyle 2.46}$
    & $44.85_{\scriptscriptstyle 5.02}$
    & $63.83_{\scriptscriptstyle 1.25}$
    & $85.07_{\scriptscriptstyle 0.37}$
    & $61.92_{\scriptscriptstyle 1.12}$
    & $36.12_{\scriptscriptstyle 5.40}$ \\

    MDE-MIL
    & $60.62_{\scriptscriptstyle 1.82}$
    & $85.83_{\scriptscriptstyle 2.56}$
    & $52.95_{\scriptscriptstyle 2.70}$
    & $17.81_{\scriptscriptstyle 4.95}$
    & $56.08_{\scriptscriptstyle 1.62}$
    & $86.79_{\scriptscriptstyle 0.97}$
    & $54.66_{\scriptscriptstyle 2.43}$
    & $45.52_{\scriptscriptstyle 6.75}$
    & $65.50_{\scriptscriptstyle 1.72}$
    & $87.24_{\scriptscriptstyle 0.50}$
    & $64.09_{\scriptscriptstyle 1.68}$
    & $39.39_{\scriptscriptstyle 4.22}$ \\

    \textbf{DeCo-MIL}
    & $\mathbf{68.44}_{\scriptscriptstyle\mathbf{1.17}}$
    & $\mathbf{91.08}_{\scriptscriptstyle\mathbf{0.36}}$
    & $\mathbf{64.65}_{\scriptscriptstyle\mathbf{1.11}}$
    & $\mathbf{36.63}_{\scriptscriptstyle\mathbf{4.13}}$
    & $\mathbf{59.82}_{\scriptscriptstyle\mathbf{0.39}}$
    & $\mathbf{88.63}_{\scriptscriptstyle\mathbf{0.20}}$
    & $\mathbf{59.49}_{\scriptscriptstyle\mathbf{0.43}}$
    & $\mathbf{53.07}_{\scriptscriptstyle\mathbf{1.34}}$
    & $\mathbf{67.17}_{\scriptscriptstyle\mathbf{3.14}}$
    & $\mathbf{88.89}_{\scriptscriptstyle\mathbf{0.54}}$
    & $\mathbf{66.69}_{\scriptscriptstyle\mathbf{3.28}}$
    & $\mathbf{47.99}_{\scriptscriptstyle\mathbf{5.68}}$ \\

    \bottomrule
  \end{tabular}%
  }
  \caption{Comparison with baseline methods on three long-tailed pathology datasets. Subscripts denote standard deviations.}
  \label{tab:baseline_comparison}
\end{table*}

\begin{figure*}[t]
\centering
\includegraphics[width=0.96\textwidth]{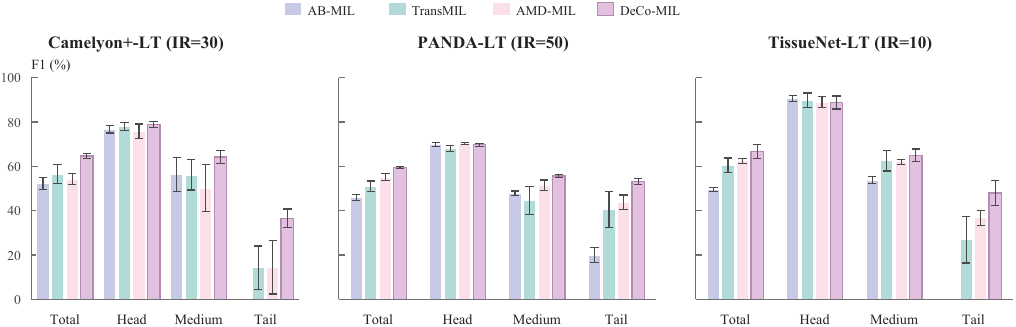} 
\caption{Overall and group-wise F1 scores of different methods on three long-tailed pathology datasets. }
\label{fig3}
\end{figure*}

\subsection{Training Objective}

The final
model is optimized using

\begin{equation}
\mathcal{L}_{\mathrm{final}}
=
\mathcal{L}_{\mathrm{main}}
+
\mathcal{L}_{\mathrm{pseudo}}
+
\beta\mathcal{L}_{\mathrm{cons}},
\label{eq:final-loss}
\end{equation}

where $\mathcal{L}_{\mathrm{main}}$ applies cross-entropy
supervision to the redundancy-reduced bag $X^*$, and
$\mathcal{L}_{\mathrm{pseudo}}$ averages the cross-entropy
losses over the $M_{\mathrm{pb}}$ pseudo-bags using their
source-slide label $y$.

To reduce the prediction variance caused by random pseudo-bag
sampling, we further encourage pseudo-bags from the same slide
to produce consistent predictions. Let $p^{(m)}$ denote the predicted distribution of pseudo-bag $B^{(m)}$ and
$\bar p=
\frac{1}{M_{\mathrm{pb}}}
\sum_{m=1}^{M_{\mathrm{pb}}}p^{(m)}$ their mean
prediction. The consistency loss is defined as:

\begin{equation}
\mathcal{L}_{\mathrm{cons}}
=
\frac{1}{M_{\mathrm{pb}}}
\sum_{m=1}^{M_{\mathrm{pb}}}
\operatorname{KL}
\left(
p^{(m)}\,\|\,\bar p
\right).
\label{eq:consistency-loss}
\end{equation}

Here, $\beta$ controls the consistency regularization and is set to $0.5$ for all datasets.
At inference, only $f_{\mathrm{final}}$ is retained and applied
to the original full bag $X$; the judge model, redundancy
masking, and pseudo-bag construction are used only during training.


\section{Experiments}
\label{sec:experiments}

\subsection{Datasets and Evaluation Metrics}
\label{sec:experimental-setup}


We evaluate DeCo-MIL on three multi-class WSI datasets: Camelyon+-LT for four-class lymph-node metastasis classification, PANDA-LT for six-class prostate ISUP grading, and TissueNet-LT for four-class cervical lesion classification. Camelyon+-LT retains its natural long-tailed distribution ($IR=30$), while PANDA-LT and TissueNet-LT are constructed with training-set imbalance ratios of 50 and 10, respectively. The validation and test sets remain class-balanced.
We report slide-level ACC, one-vs-rest macro-AUC, macro-F1, and frequency-group F1 averaged as the mean and standard deviation over five independent runs. Detailed data statistics and
group assignments are provided in the supplementary material.

\subsection{Implementation Details}
\label{sec:implementation-details}

All experiments are implemented in PyTorch and conducted on a workstation equipped with an NVIDIA A100 GPU (80 GB). Each WSI is tiled at $20\times$ magnification into non-overlapping $512\times512  $ patches. Patch features are extracted using the pretrained pathology foundation model CONCH~\cite{conch} and then $\ell_2$-normalized. In DeCo-MIL, the judge model $f_{\mathrm{judge}}$ and final model $f_{\mathrm{final}}$ both use AB-MIL as the default MIL backbone.

Uniform hyperparameters are maintained across all experiments. Models are optimized using Adam with an initial learning rate of $2\times10^{-4}$ and a weight decay of $1\times10^{-5}$. The checkpoint with the highest validation macro-F1 is selected.

\subsection{Comparison with State-of-the-Art Methods}
\label{sec:main-comparison}

We compare DeCo-MIL with eleven representative WSI MIL methods:
AB-MIL~\cite{abmil}, CLAM-SB/MB~\cite{clam},
DS-MIL~\cite{dsmil}, TransMIL~\cite{transmil},
WiKG~\cite{wikg}, ILRA-MIL~\cite{ilra},
MHIM-MIL~\cite{mhim}, AMD-MIL~\cite{amdmil},
MiCo~\cite{micomil}, and MDE-MIL~\cite{mdemil}.

 As shown in Table~\ref{tab:baseline_comparison}, DeCo-MIL achieves the highest mean on every reported metric across the three datasets, with the most pronounced gains on tail classes. 
Compared with the best-performing Tail-F1 baseline on each dataset, DeCo-MIL achieves absolute gains of $18.82$, $7.40$, and
$6.98$ percentage points on Camelyon+-LT, PANDA-LT, and
TissueNet-LT, respectively, demonstrating that its debiased counterfactual reasoning effectively captures scarce tail-class evidence and enhances rare-class recognition under the nested dual long-tailed setting of WSI classification.

As shown in Fig.~\ref{fig3}, DeCo-MIL substantially improves Tail-F1 while maintaining Head-F1 comparable to strong baselines on all three datasets; it also improves Medium-F1. The tail-class gains are therefore not obtained at the expense of head-class recognition.

\begin{table}[t]
  \centering

  \setlength{\tabcolsep}{2.0pt}
  \renewcommand{\arraystretch}{0.96}

  \resizebox{\columnwidth}{!}{%
  \begin{tabular}{llcccc}
    \toprule
    Dataset & Module & ACC(\%)  & AUC(\%)  & F1(\%)  & Tail-F1(\%)  \\
    \midrule

    \multirow{5}{*}{Camelyon+}
    & Base
    & $61.56_{\scriptscriptstyle 2.54}$
    & $87.76_{\scriptscriptstyle 1.69}$
    & $52.46_{\scriptscriptstyle 2.71}$
    & $0.00_{\scriptscriptstyle 0.00}$ \\

    & + Counterfactual
    & $65.94_{\scriptscriptstyle 1.17}$
    & $85.27_{\scriptscriptstyle 0.53}$
    & $59.45_{\scriptscriptstyle 1.60}$
    & $13.86_{\scriptscriptstyle 4.18}$ \\

    & + Debiasing
    & $64.38_{\scriptscriptstyle 1.82}$
    & $89.67_{\scriptscriptstyle 0.86}$
    & $58.18_{\scriptscriptstyle 1.85}$
    & $18.04_{\scriptscriptstyle 5.12}$ \\

    & + Pseudo-bag
    & $63.75_{\scriptscriptstyle 1.82}$
    & $90.18_{\scriptscriptstyle 1.19}$
    & $58.96_{\scriptscriptstyle 1.72}$
    & $29.16_{\scriptscriptstyle 4.10}$ \\

    & + Oversampling
    & $\mathbf{68.44}_{\scriptscriptstyle\mathbf{1.17}}$
    & $\mathbf{91.08}_{\scriptscriptstyle\mathbf{0.36}}$
    & $\mathbf{64.65}_{\scriptscriptstyle\mathbf{1.11}}$
    & $\mathbf{36.63}_{\scriptscriptstyle\mathbf{4.13}}$ \\

    \midrule

    \multirow{5}{*}{PANDA}
    & Base
    & $51.28_{\scriptscriptstyle 0.83}$
    & $86.74_{\scriptscriptstyle 0.36}$
    & $45.90_{\scriptscriptstyle 1.30}$
    & $19.85_{\scriptscriptstyle 3.38}$ \\

    & + Counterfactual
    & $57.30_{\scriptscriptstyle 1.05}$
    & $88.26_{\scriptscriptstyle 0.14}$
    & $56.07_{\scriptscriptstyle 1.35}$
    & $45.36_{\scriptscriptstyle 2.55}$ \\

    & + Debiasing
    & $58.18_{\scriptscriptstyle 0.72}$
    & $88.60_{\scriptscriptstyle 0.14}$
    & $56.74_{\scriptscriptstyle 1.21}$
    & $45.03_{\scriptscriptstyle 3.92}$ \\

    & + Pseudo-bag
    & $59.62_{\scriptscriptstyle 0.45}$
    & $\mathbf{89.03}_{\scriptscriptstyle\mathbf{0.18}}$
    & $58.48_{\scriptscriptstyle 0.71}$
    & $48.27_{\scriptscriptstyle 2.77}$ \\

    & + Oversampling
    & $\mathbf{59.82}_{\scriptscriptstyle\mathbf{0.39}}$
    & $88.63_{\scriptscriptstyle 0.20}$
    & $\mathbf{59.49}_{\scriptscriptstyle\mathbf{0.43}}$
    & $\mathbf{53.07}_{\scriptscriptstyle\mathbf{1.34}}$ \\

    \midrule

    \multirow{5}{*}{TissueNet}
    & Base
    & $56.67_{\scriptscriptstyle 1.05}$
    & $83.06_{\scriptscriptstyle 0.96}$
    & $49.64_{\scriptscriptstyle 0.81}$
    & $0.00_{\scriptscriptstyle 0.00}$ \\

    & + Counterfactual
    & $58.33_{\scriptscriptstyle 4.12}$
    & $84.00_{\scriptscriptstyle 1.33}$
    & $56.48_{\scriptscriptstyle 4.87}$
    & $33.12_{\scriptscriptstyle 7.42}$ \\

    & + Debiasing
    & $60.00_{\scriptscriptstyle 1.58}$
    & $83.90_{\scriptscriptstyle 0.54}$
    & $58.90_{\scriptscriptstyle 1.75}$
    & $36.29_{\scriptscriptstyle 4.72}$ \\

    & + Pseudo-bag
    & $62.50_{\scriptscriptstyle 2.64}$
    & $85.34_{\scriptscriptstyle 0.66}$
    & $61.46_{\scriptscriptstyle 2.33}$
    & $39.11_{\scriptscriptstyle 3.41}$ \\

    & + Oversampling
    & $\mathbf{67.17}_{\scriptscriptstyle\mathbf{3.14}}$
    & $\mathbf{88.89}_{\scriptscriptstyle\mathbf{0.54}}$
    & $\mathbf{66.69}_{\scriptscriptstyle\mathbf{3.28}}$
    & $\mathbf{47.99}_{\scriptscriptstyle\mathbf{5.68}}$ \\

    \bottomrule
  \end{tabular}%
  }
  \caption{Component ablation on three long-tailed pathology datasets.}
  \label{tab:component_ablation}
  
\end{table}

\subsection{Ablation Studies}
\label{sec:ablation}

\subsubsection{Component Ablation}
\label{sec:component-ablation}


Table~\ref{tab:component_ablation} evaluates the incremental
contribution of each component. 
Counterfactual-guided masking provides the most immediate gains for tail classes,
demonstrating the importance of removing redundant instances while preserving discriminative evidence. Frequency debiasing further improves tail-class performance on
Camelyon+-LT and TissueNet-LT, while improving the overall metrics on PANDA-LT
with its Tail-F1 largely preserved, demonstrating that class-frequency calibration
reduces interference from head-class priors.
Finally, sequentially adding stratified pseudo-bags and
tail-aware oversampling consistently improves Tail-F1 across
all three datasets, validating their complementary roles in
preserving morphology coverage and strengthening effective
supervision for tail classes.
Overall, all proposed components play key roles in DeCo-MIL
and jointly mitigate the nested dual long-tail problem.

\subsubsection{Effectiveness of Debiased Counterfactual-Guided Masking}
\label{sec:masking-ablation}

\begin{table}[t]
  \centering
  \setlength{\tabcolsep}{1.5pt}
  \renewcommand{\arraystretch}{1.10}
  \resizebox{\columnwidth}{!}{%
  \begin{tabular}{lcccccc}
    \toprule
    \multirow{2}{*}{Strategy} & \multicolumn{2}{c}{Camelyon+-LT} & \multicolumn{2}{c}{PANDA-LT} & \multicolumn{2}{c}{TissueNet-LT} \\
    \cmidrule(lr){2-3} \cmidrule(lr){4-5} \cmidrule(lr){6-7}
    & F1(\%) & Tail-F1(\%) & F1(\%) & Tail-F1(\%) & F1(\%) & Tail-F1(\%) \\
    \midrule
    Random & $56.21_{\scriptscriptstyle 2.25}$ & $25.09_{\scriptscriptstyle 7.54}$ & $57.94_{\scriptscriptstyle 1.21}$ & $48.37_{\scriptscriptstyle 1.82}$ & $63.20_{\scriptscriptstyle 0.59}$ & $38.57_{\scriptscriptstyle 2.18}$ \\
    Attention & $58.16_{\scriptscriptstyle 1.51}$ & $20.34_{\scriptscriptstyle 7.52}$ & $57.48_{\scriptscriptstyle 1.01}$ & $46.81_{\scriptscriptstyle 1.43}$ & $63.81_{\scriptscriptstyle 2.56}$ & $41.91_{\scriptscriptstyle 3.28}$ \\
    Anchor-size & $61.30_{\scriptscriptstyle 3.46}$ & $23.56_{\scriptscriptstyle 7.05}$ & $58.83_{\scriptscriptstyle 0.71}$ & $51.12_{\scriptscriptstyle 1.84}$ & $65.46_{\scriptscriptstyle 1.89}$ & $43.02_{\scriptscriptstyle 6.35}$ \\
    \textbf{Counterfactual} & $\mathbf{64.65}_{\scriptscriptstyle\mathbf{1.11}}$ & $\mathbf{36.63}_{\scriptscriptstyle\mathbf{4.13}}$ & $\mathbf{59.49}_{\scriptscriptstyle\mathbf{0.43}}$ & $\mathbf{53.07}_{\scriptscriptstyle\mathbf{1.34}}$ & $\mathbf{66.69}_{\scriptscriptstyle\mathbf{3.28}}$ & $\mathbf{47.99}_{\scriptscriptstyle\mathbf{5.68}}$ \\
    \bottomrule
  \end{tabular}%
  }
  \caption{Comparison of different masking strategies on three long-tailed pathology datasets.}
  \label{tab:mask_strategy}
\end{table}

To evaluate the role of counterfactual contributions in instance selection, we compare the proposed counterfactual-guided masking with random, attention-based, and anchor-size masking under the same maximum masking ratio and training protocol; detailed definitions are provided in the supplementary material. As shown in Table~\ref{tab:mask_strategy}, our strategy achieves the best macro-F1 and Tail-F1 across all three datasets. Random masking ignores the discriminative relevance of individual instances, while attention-based masking does not explicitly correct for class-frequency bias. Anchor-size masking favors larger tissue regions but cannot distinguish discriminative morphologies from redundant ones. In contrast, our strategy uses frequency-debiased counterfactual contributions to determine both masking targets and ratios, thereby removing redundant instances while preserving scarce tail-class evidence.


\begin{table}[t]
  \centering
  \setlength{\tabcolsep}{1.5pt}
  \renewcommand{\arraystretch}{1.10}
  \resizebox{\columnwidth}{!}{%
  \begin{tabular}{lcccccc}
    \toprule
    \multirow{2}{*}{Strategy} & \multicolumn{2}{c}{Camelyon+-LT} & \multicolumn{2}{c}{PANDA-LT} & \multicolumn{2}{c}{TissueNet-LT} \\
    \cmidrule(lr){2-3} \cmidrule(lr){4-5} \cmidrule(lr){6-7}
    & F1(\%) & Tail-F1(\%) & F1(\%) & Tail-F1(\%) & F1(\%) & Tail-F1(\%) \\
    \midrule
    Zero & $56.16_{\scriptscriptstyle 2.60}$ & $21.30_{\scriptscriptstyle 13.13}$ & $57.61_{\scriptscriptstyle 0.92}$ & $48.26_{\scriptscriptstyle 3.24}$ & $64.21_{\scriptscriptstyle 2.11}$ & $41.01_{\scriptscriptstyle 4.50}$ \\
    Anchor Proto. & $61.22_{\scriptscriptstyle 3.41}$ & $25.29_{\scriptscriptstyle 9.37}$ & $58.28_{\scriptscriptstyle 1.11}$ & $51.90_{\scriptscriptstyle 1.20}$ & $64.53_{\scriptscriptstyle 2.64}$ & $43.93_{\scriptscriptstyle 4.84}$ \\
    \textbf{Normal Proto.} & $\mathbf{64.65}_{\scriptscriptstyle\mathbf{1.11}}$ & $\mathbf{36.63}_{\scriptscriptstyle\mathbf{4.13}}$ & $\mathbf{59.49}_{\scriptscriptstyle\mathbf{0.43}}$ & $\mathbf{53.07}_{\scriptscriptstyle\mathbf{1.34}}$ & $\mathbf{66.69}_{\scriptscriptstyle\mathbf{3.28}}$ & $\mathbf{47.99}_{\scriptscriptstyle\mathbf{5.68}}$ \\
    \bottomrule
  \end{tabular}%
  }
  \caption{Comparison of different counterfactual constructions on three long-tailed pathology datasets.}
  \label{tab:counterfactual}
\end{table}

\begin{table}[t]
  \centering
  \setlength{\tabcolsep}{1.5pt}
  \renewcommand{\arraystretch}{1.10}
  \resizebox{\columnwidth}{!}{%
  \begin{tabular}{lcccccc}
    \toprule
    \multirow{2}{*}{Strategy} & \multicolumn{2}{c}{Camelyon+-LT} & \multicolumn{2}{c}{PANDA-LT} & \multicolumn{2}{c}{TissueNet-LT} \\
    \cmidrule(lr){2-3} \cmidrule(lr){4-5} \cmidrule(lr){6-7}
    & F1(\%) & Tail-F1(\%) & F1(\%) & Tail-F1(\%) & F1(\%) & Tail-F1(\%) \\
    \midrule
    Random & $58.20_{\scriptscriptstyle 3.44}$ & $13.11_{\scriptscriptstyle 12.41}$ & $57.19_{\scriptscriptstyle 0.54}$ & $47.51_{\scriptscriptstyle 1.09}$ & $64.35_{\scriptscriptstyle 1.82}$ & $38.48_{\scriptscriptstyle 4.49}$ \\
    Equal-per-Anchor & $61.31_{\scriptscriptstyle 4.27}$ & $22.72_{\scriptscriptstyle 14.89}$ & $58.17_{\scriptscriptstyle 0.48}$ & $50.47_{\scriptscriptstyle 1.56}$ & $64.26_{\scriptscriptstyle 2.43}$ & $42.17_{\scriptscriptstyle 3.20}$ \\
    \textbf{Stratified} & $\mathbf{64.65}_{\scriptscriptstyle\mathbf{1.11}}$ & $\mathbf{36.63}_{\scriptscriptstyle\mathbf{4.13}}$ & $\mathbf{59.49}_{\scriptscriptstyle\mathbf{0.43}}$ & $\mathbf{53.07}_{\scriptscriptstyle\mathbf{1.34}}$ & $\mathbf{66.69}_{\scriptscriptstyle\mathbf{3.28}}$ & $\mathbf{47.99}_{\scriptscriptstyle\mathbf{5.68}}$ \\
    \bottomrule
  \end{tabular}%
  }
  \caption{Effectiveness of different sampling strategies on three long-tailed pathology datasets.}
  \label{tab:sampling_strategy}
\end{table}

We further compare three counterfactual constructions:
replacing all instances assigned to the target anchor with
a zero vector, the corresponding morphology prototype $H_k$,
or the matched normal prototype $\hat q_k$ used in our method.
As shown in Table~\ref{tab:counterfactual}, matched
normal prototype replacement performs best across all three
datasets. Zero replacement produces off-manifold
interventional features, whereas $H_k$ retains substantial
anchor-specific tissue semantics. In contrast, $\hat q_k$
attenuates pathology-specific information while remaining
closer to the distribution of normal features, making it a more
suitable counterfactual reference. Together, these results
support the effectiveness of frequency-debiased
counterfactual-guided masking and matched normal prototype
intervention.

\subsubsection{Effectiveness of Anchor-Stratified Pseudo-Bag Sampling}
\label{sec:pseudobag-ablation}

To evaluate the pseudo-bag construction strategy, we compare
random, equal-per-anchor, and the proposed anchor-stratified
sampling under the same sampling ratio and number of
pseudo-bags. As shown in Table~\ref{tab:sampling_strategy}, stratified sampling achieves the highest macro-F1 and Tail-F1 across all three datasets, demonstrating consistent gains on tail classes. Random sampling can omit scarce discriminative regions, whereas equal-per-anchor sampling alters the relative composition across different tissue morphologies. The proposed strategy samples instances proportionally from each retained anchor, preserving both tissue coverage and relative instance composition while making fuller use of the limited
discriminative evidence in tail-class slides.

\begin{figure}[t]
\centering
\includegraphics[width=0.95\columnwidth]{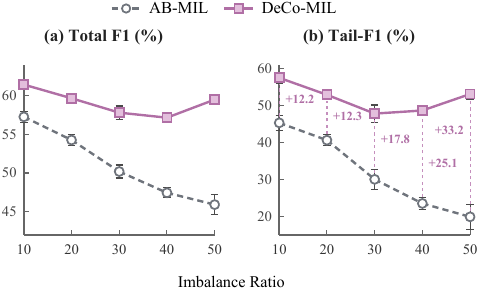} 
\caption{Performance comparison under different imbalance ratios on PANDA-LT. (a) Total F1 and (b) Tail-F1 of AB-MIL and DeCo-MIL. The purple annotations denote the Tail-F1 percentage-point gains of DeCo-MIL over AB-MIL.}
\label{ir_robustness}
\end{figure}

\subsection{Robustness to Varying Imbalance Ratios}
\label{sec:ir-robustness}

To evaluate the robustness of DeCo-MIL under different degrees of long-tailed imbalance, we vary the training-set imbalance ratio on PANDA-LT as $IR\in\{10,20,30,40,50\}$ and compare DeCo-MIL with AB-MIL on the same balanced test set. As shown in Fig.~\ref{ir_robustness}, the overall and tail-class performance of AB-MIL declines markedly as $IR$ increases, with particularly severe degradation in Tail-F1. In contrast, DeCo-MIL maintains more stable overall and tail-class performance across different imbalance ratios, while its advantage consistently widens as the imbalance becomes more severe. These results show that DeCo-MIL effectively mitigates tail-class performance degradation caused by long-tailed distributions and is better suited to severely imbalanced settings.

\begin{figure}[t]
\centering
\includegraphics[width=1.0\columnwidth]{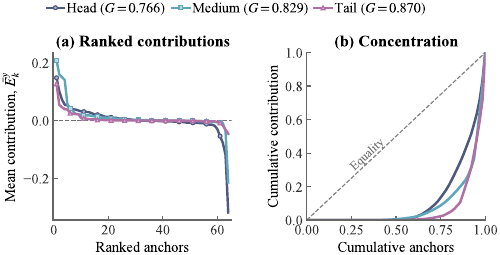} 
\caption{Nested dual long-tail analysis on PANDA-LT ($IR=50$): 
(a) ranked debiased counterfactual contributions of tissue anchors and 
(b) the corresponding Lorenz curves and Gini coefficients.}
\label{instance_longtail}
\end{figure}

\subsection{Visualization and Analysis}
\label{sec:visualization-analysis}

\subsubsection{Analysis of the Nested Dual Long-Tail}
\label{sec:nested-long-tail-analysis}

We aggregate the debiased counterfactual contributions $\bar{E}_k^y$ produced by the judge on PANDA-LT ($IR=50$) according to the Head, Medium, and Tail groups. As shown in Fig.~\ref{instance_longtail}(a), only a few anchors in each group carry the dominant discriminative information, while the remaining contributions are close to zero or negative. After clipping negative contributions to zero, the Gini coefficient in Fig.~\ref{instance_longtail}(b) increases monotonically from Head to Tail. Moreover, the top 10\% of anchors account for 53\%--74\% of the total positive contribution while covering only 15\%--22\% of the patches. These results show that discriminative evidence is concentrated in a small number of tissue morphologies and becomes increasingly concentrated for rarer classes, providing empirical support for the coupling between the class-level long tail and instance-level evidence long tail.

\subsubsection{Interpretability Analysis}

\begin{figure}[t]
\centering
\includegraphics[width=1.0\columnwidth]{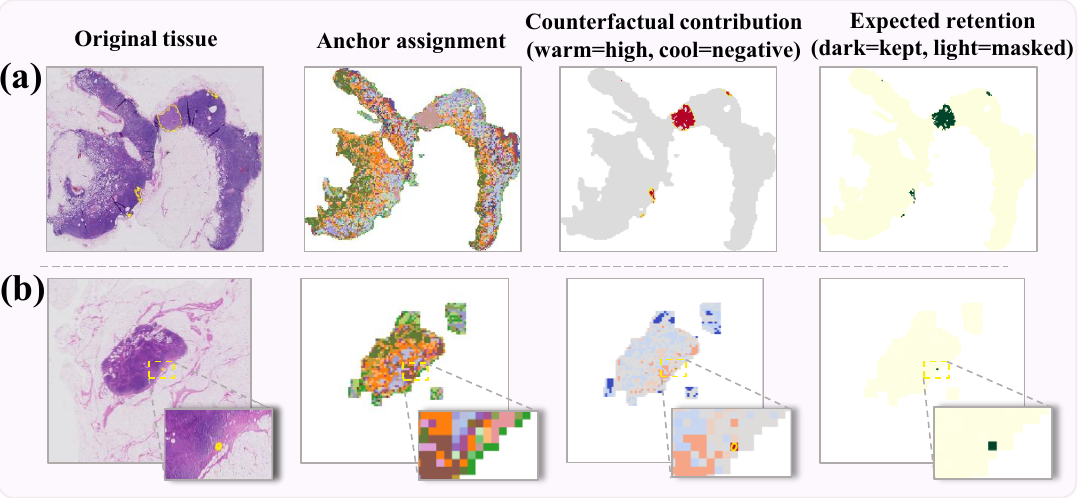} 
\caption{Visualization of DeCo-MIL on tail-class
slides from Camelyon+-LT: From left to right: the original WSI with official lesion annotations, morphology-anchor assignments, debiased counterfactual contribution heatmaps, and expected retention maps. Yellow contours denote the official lesion annotations.}
\label{interpretability}
\end{figure}

We conduct an interpretability analysis of DeCo-MIL using tail-class slides from Camelyon+-LT, as shown in
Fig.~\ref{interpretability}. Regions with high positive
contributions exhibit clear spatial agreement with the annotated lesions and
consequently receive higher retention probabilities, while most
non-discriminative regions are preferentially masked. These results visually
illustrate that debiased counterfactual reasoning can identify lesion-related
tissue morphologies and guide redundancy masking to preserve scarce
discriminative evidence for tail classes.

\section{Conclusion}

This paper studies long-tailed WSI analysis from the perspective of a
nested dual long-tail, where the inter-slide class long tail is coupled with
the intra-slide long tail of instance-level discriminative evidence. We
propose DeCo-MIL, which uses frequency-debiased counterfactual reasoning with
matched normal prototypes to quantify the discriminative contributions of
tissue-morphology anchors and guide redundancy masking, thereby preserving
scarce discriminative instances. It further combines anchor-stratified
pseudo-bags, tail-aware oversampling, and consistency regularization to
strengthen supervision for tail classes while maintaining relative
tissue-morphology composition. Experiments on three long-tailed WSI datasets show that DeCo-MIL achieves state-of-the-art overall and tail-class performance. Component
ablations, robustness evaluations across imbalance ratios, and visualization
analyses further validate the proposed designs and provide empirical support
for the nested dual long-tail.

\bibliography{aaai2027}

@inproceedings{abmil,
  title     = {Attention-based Deep Multiple Instance Learning},
  author    = {Ilse, Maximilian and Tomczak, Jakub M. and Welling, Max},
  booktitle = {International Conference on Machine Learning (ICML)},
  pages     = {2127--2136},
  year      = {2018}
}

@article{clam,
  title   = {Data-efficient and weakly supervised computational pathology on whole-slide images},
  author  = {Lu, Ming Y. and Williamson, Drew F. K. and Chen, Tiffany Y. and Chen, Richard J. and Barbieri, Matteo and Mahmood, Faisal},
  journal = {Nature Biomedical Engineering},
  volume  = {5},
  number  = {6},
  pages   = {555--570},
  year    = {2021}
}

@inproceedings{transmil,
  title     = {{TransMIL}: Transformer Based Correlated Multiple Instance Learning for Whole Slide Image Classification},
  author    = {Shao, Zhuchen and Bian, Hao and Chen, Yang and Wang, Yifeng and Zhang, Jian and Ji, Xiangyang and Zhang, Yongbing},
  booktitle = {Advances in Neural Information Processing Systems},
  volume    = {34},
  pages     = {2136--2147},
  year      = {2021}
}

@inproceedings{dsmil,
  title     = {Dual-stream Multiple Instance Learning Network for Whole Slide Image Classification with Self-supervised Contrastive Learning},
  author    = {Li, Bin and Li, Yin and Eliceiri, Kevin W.},
  booktitle = {IEEE/CVF Conference on Computer Vision and Pattern Recognition (CVPR)},
  pages     = {14318--14328},
  year      = {2021}
}

@inproceedings{mhim,
  title     = {Multiple Instance Learning Framework with Masked Hard Instance Mining for Whole Slide Image Classification},
  author    = {Tang, Wenhao and Huang, Sheng and Zhang, Xiaoxian and Zhou, Fengtao and Zhang, Yi and Liu, Bo},
  booktitle = {Proceedings of the IEEE/CVF International Conference on Computer Vision},
  pages     = {4078--4087},
  year      = {2023}
}

@ARTICLE{frmil,
  author = {Chikontwe, Philip and
          Kim, Meejeong and
          Jeong, Jaehoon and
          Sung, Hyun Jung and
          Go, Heounjeong and
          Nam, Soo Jeong and
          Park, Sang Hyun},
  journal={IEEE Transactions on Medical Imaging}, 
  title={FR-MIL: Distribution Re-Calibration-Based Multiple Instance Learning With Transformer for Whole Slide Image Classification}, 
  year={2025},
  volume={44},
  number={1},
  pages={409-421},
  doi={10.1109/TMI.2024.3446716}}

@inproceedings{amdmil,
  title={Agent aggregator with mask denoise mechanism for histopathology whole slide image analysis},
  author={Ling, Xitong and Ouyang, Minxi and Wang, Yizhi and Chen, Xinrui and Yan, Renao and Chu, Hongbo and Cheng, Junru and Guan, Tian and Tian, Sufang and Liu, Xiaoping and others},
  booktitle={Proceedings of the 32nd ACM International Conference on Multimedia},
  pages={2795--2803},
  year={2024}
}

@inproceedings{wikg,
  title={Dynamic graph representation with knowledge-aware attention for histopathology whole slide image analysis},
  author={Li, Jiawen and Chen, Yuxuan and Chu, Hongbo and Sun, Qiehe and Guan, Tian and Han, Anjia and He, Yonghong},
  booktitle={Proceedings of the IEEE/CVF conference on computer vision and pattern recognition},
  pages={11323--11332},
  year={2024}
}

@inproceedings{h2mil,
  title     = {{H$^2$-MIL}: Exploring Hierarchical Representation with Heterogeneous Multiple Instance Learning for Whole Slide Image Analysis},
  author    = {Hou, Wentai and Yu, Lequan and Lin, Chengxuan and Huang, Helong and Yu, Rongshan and Qin, Jing and Wang, Liansheng},
  booktitle = {Proceedings of the AAAI Conference on Artificial Intelligence},
  volume    = {36},
  number    = {1},
  pages     = {933--941},
  year      = {2022},
  doi       = {10.1609/aaai.v36i1.19976}
}

@inproceedings{micomil,
  title={MiCo: Multiple Instance Learning with Context-Aware Clustering for Whole Slide Image Analysis},
  author={Li, Junjian and Liu, Jin and Kuang, Hulin and Yue, Hailin and He, Mengshen and Wang, Jianxin},
  booktitle={International Conference on Medical Image Computing and Computer-Assisted Intervention},
  pages={376--385},
  year={2025},
  organization={Springer}
}

@inproceedings{pamoe,
  title={Learning heterogeneous tissues with mixture of experts for gigapixel whole slide images},
  author={Wu, Junxian and Chen, Minheng and Ke, Xinyi and Xun, Tianwang and Jiang, Xiaoming and Zhou, Hongyu and Shao, Lizhi and Kong, Youyong},
  booktitle={Proceedings of the Computer Vision and Pattern Recognition Conference},
  pages={5144--5153},
  year={2025}
}

@inproceedings{panther,
  title={Morphological prototyping for unsupervised slide representation learning in computational pathology},
  author={Song, Andrew H and Chen, Richard J and Ding, Tong and Williamson, Drew FK and Jaume, Guillaume and Mahmood, Faisal},
  booktitle={Proceedings of the IEEE/CVF Conference on Computer Vision and Pattern Recognition},
  pages={11566--11578},
  year={2024}
}

@inproceedings{ilra,
  title     = {Exploring Low-Rank Property in Multiple Instance Learning for Whole Slide Image Classification},
  author    = {Xiang, Jinxi and Wang, Xiyue and Zhang, Jun and Yang, Sen and Han, Xiao and Yang, Wei},
  booktitle = {The Eleventh International Conference on Learning Representations},
  year      = {2023}
}

@inproceedings{sammil,
  title     = {{SAM-MIL}: A Spatial Contextual Aware Multiple Instance Learning Approach for Whole Slide Image Classification},
  author    = {Fang, Heng and Huang, Sheng and Tang, Wenhao and Huangfu, Luwen and Liu, Bo},
  booktitle = {Proceedings of the ACM International Conference on Multimedia (ACM MM)},
  pages     = {6083--6092},
  year      = {2024}
}

@inproceedings{mdemil,
  title={Multimodal Distillation-Driven Ensemble Learning for Long-Tailed Histopathology Whole Slide Images Analysis},
  author={Ling, Xitong and Li, Xiaoxiao and Li, Jiawen and Fu, Mingxi and Chen, Yuxuan and Peng, Jing and Ouyang, Minxi and Wang, Yizhi and He, Yonghong and Liu, Xiaoping and others},
  booktitle={2026 IEEE 23rd International Symposium on Biomedical Imaging (ISBI)},
  pages={1--5},
  year={2026},
  organization={IEEE}
}

@InProceedings{scmil,
    author    = {Juyal, Dinkar and Shingi, Siddhant and Javed, Syed Ashar and Padigela, Harshith and Shah, Chintan and Sampat, Anand and Khosla, Archit and Abel, John and Taylor-Weiner, Amaro},
    title     = {SC-MIL: Supervised Contrastive Multiple Instance Learning for Imbalanced Classification in Pathology},
    booktitle = {Proceedings of the IEEE/CVF Winter Conference on Applications of Computer Vision (WACV)},
    month     = {January},
    year      = {2024},
    pages     = {7946-7955}
}

@inproceedings{shap,
  title     = {A Unified Approach to Interpreting Model Predictions},
  author    = {Lundberg, Scott M. and Lee, Su-In},
  booktitle = {Advances in Neural Information Processing Systems},
  volume    = {30},
  pages     = {4765--4774},
  year      = {2017}
}

@inproceedings{cal,
  title     = {Counterfactual Attention Learning for Fine-Grained Visual Categorization and Re-Identification},
  author    = {Rao, Yongming and Chen, Guangyi and Lu, Jiwen and Zhou, Jie},
  booktitle = {Proceedings of the IEEE/CVF International Conference on Computer Vision},
  pages     = {1025--1034},
  year      = {2021}
}

@inproceedings{deconfoundtde,
  title     = {Long-Tailed Classification by Keeping the Good and Removing the Bad Momentum Causal Effect},
  author    = {Tang, Kaihua and Huang, Jianqiang and Zhang, Hanwang},
  booktitle = {Advances in Neural Information Processing Systems},
  volume    = {33},
  pages     = {1513--1524},
  year      = {2020}
}

@inproceedings{cimil,
  title     = {Boosting Multiple Instance Learning Models for Whole Slide Image Classification: A Model-Agnostic Framework Based on Counterfactual Inference},
  author    = {Lin, Weiping and Zhuang, Zhenfeng and Yu, Lequan and Wang, Liansheng},
  booktitle = {Proceedings of the AAAI Conference on Artificial Intelligence},
  pages     = {3477--3485},
  year      = {2024}
}

@inproceedings{ibmil,
  title     = {Interventional Bag Multi-Instance Learning on Whole-Slide Pathological Images},
  author    = {Lin, Tiancheng and Yu, Zhimiao and Hu, Hongyu and Xu, Yi and Chen, Chang-Wen},
  booktitle = {IEEE/CVF Conference on Computer Vision and Pattern Recognition (CVPR)},
  pages     = {19830--19839},
  year      = {2023}
}

@article{conch,
  title   = {A Visual-Language Foundation Model for Computational Pathology},
  author  = {Lu, Ming Y. and Chen, Bowen and Williamson, Drew F. K. and
             Chen, Richard J. and Liang, Ivy and Ding, Tong and
             Jaume, Guillaume and Odintsov, Igor and Le, Long Phi and
             Gerber, Georg and Parwani, Anil V. and Zhang, Andrew and
             Mahmood, Faisal},
  journal = {Nature Medicine},
  volume  = {30},
  number  = {3},
  pages   = {863--874},
  year    = {2024},
  doi     = {10.1038/s41591-024-02856-4}
}

@inproceedings{zeiler2014,
  title     = {Visualizing and Understanding Convolutional Networks},
  author    = {Zeiler, Matthew D. and Fergus, Rob},
  booktitle = {European Conference on Computer Vision (ECCV)},
  pages     = {818--833},
  year      = {2014},
  publisher = {Springer}
}

@inproceedings{fong2017,
  title     = {Interpretable Explanations of Black Boxes by Meaningful Perturbation},
  author    = {Fong, Ruth C. and Vedaldi, Andrea},
  booktitle = {IEEE International Conference on Computer Vision (ICCV)},
  pages     = {3429--3437},
  year      = {2017}
}

@inproceedings{cao2019ldam,
  title     = {Learning Imbalanced Datasets with
               Label-Distribution-Aware Margin Loss},
  author    = {Cao, Kaidi and Wei, Colin and Gaidon, Adrien
               and Arechiga, Nikos and Ma, Tengyu},
  booktitle = {Advances in Neural Information Processing Systems},
  volume    = {32},
  pages     = {1565--1576},
  year      = {2019}
}

@inproceedings{chou2020remix,
  title     = {Remix: Rebalanced Mixup},
  author    = {Chou, Hsin-Ping and Chang, Shih-Chieh and Pan, Jia-Yu
               and Wei, Wei and Juan, Da-Cheng},
  booktitle = {Computer Vision -- ECCV 2020 Workshops},
  pages     = {95--110},
  publisher = {Springer},
  year      = {2020},
  doi       = {10.1007/978-3-030-65414-6_9}
}

@inproceedings{ren2020balancedsoftmax,
  title     = {Balanced Meta-Softmax for Long-Tailed Visual Recognition},
  author    = {Ren, Jiawei and Yu, Cunjun and Sheng, Shunan and Ma, Xiao
               and Zhao, Haiyu and Yi, Shuai and Li, Hongsheng},
  booktitle = {Advances in Neural Information Processing Systems},
  volume    = {33},
  pages     = {4175--4186},
  year      = {2020}
}

@inproceedings{menon2021logit,
  title     = {Long-Tail Learning via Logit Adjustment},
  author    = {Menon, Aditya Krishna and Jayasumana, Sadeep
               and Rawat, Ankit Singh and Jain, Himanshu
               and Veit, Andreas and Kumar, Sanjiv},
  booktitle = {International Conference on Learning Representations},
  year      = {2021}
}

@inproceedings{wang2017modeltail,
  title     = {Learning to Model the Tail},
  author    = {Wang, Yu-Xiong and Ramanan, Deva and Hebert, Martial},
  booktitle = {Advances in Neural Information Processing Systems},
  volume    = {30},
  pages     = {7029--7039},
  year      = {2017}
}

@inproceedings{kang2020decoupling,
  title     = {Decoupling Representation and Classifier for Long-Tailed Recognition},
  author    = {Kang, Bingyi and Xie, Saining and Rohrbach, Marcus and Yan, Zhicheng
               and Gordo, Albert and Feng, Jiashi and Kalantidis, Yannis},
  booktitle = {International Conference on Learning Representations},
  year      = {2020}
}

@inproceedings{wang2021ride,
  title     = {Long-Tailed Recognition by Routing Diverse Distribution-Aware Experts},
  author    = {Wang, Xudong and Lian, Long and Miao, Zhongqi and Liu, Ziwei
               and Yu, Stella X.},
  booktitle = {International Conference on Learning Representations},
  year      = {2021}
}

@article{buda2018systematic,
  title   = {A Systematic Study of the Class Imbalance Problem
             in Convolutional Neural Networks},
  author  = {Buda, Mateusz and Maki, Atsuto and Mazurowski, Maciej A.},
  journal = {Neural Networks},
  volume  = {106},
  pages   = {249--259},
  year    = {2018},
  doi     = {10.1016/j.neunet.2018.07.011}
}

\clearpage

\urlstyle{rm}
\def\UrlFont{\rm}
\frenchspacing

\pdfinfo{
/TemplateVersion (2027.1)
}

\setcounter{secnumdepth}{2}
\numberwithin{equation}{section}
\numberwithin{table}{section}
\numberwithin{figure}{section}

\newcommand{\ablationplaceholder}[1]{%
  \fbox{\parbox[c][0.13\textheight][c]{0.94\textwidth}{\centering\itshape #1}}%
}

\title{Supplementary Material for DeCo-MIL: Debiased Counterfactual Reasoning for Long-Tailed Whole Slide Image Analysis}
\author{Anonymous Authors}
\affiliations{Anonymous Institution}
\setcounter{dbltopnumber}{3}
\renewcommand{\dbltopfraction}{0.98}
\renewcommand{\dblfloatpagefraction}{0.90}

\maketitle
\appendix

\section{Dataset Details and Long-Tail Construction}

\subsection{Common Preprocessing and Split Protocol}

We evaluate DeCo-MIL on three multi-class whole-slide image (WSI) datasets: naturally long-tailed Camelyon+-LT and artificially long-tailed PANDA-LT and TissueNet-LT. WSIs are tiled at $20\times$ magnification into non-overlapping $512\times512$ patches. Patch features are extracted with the pretrained CONCH pathology foundation model and $\ell_2$-normalized before prototype construction and MIL training. The frozen feature encoder is shared by all methods.

For each class, slides are independently shuffled to reserve fixed class-balanced quotas for validation and testing, with the remaining slides forming the training pool. The natural training distribution is retained for Camelyon+-LT. For PANDA-LT and TissueNet-LT, long-tailed training sets are constructed by class-wise downsampling from the training pool. Validation and test sets remain class-balanced and are never downsampled.

For an artificially long-tailed training set, the imbalance ratio is
\begin{equation}
\mathrm{IR}=\frac{n_{\max}}{n_{\min}},
\end{equation}
where $n_{\max}$ and $n_{\min}$ are the numbers of training slides in the head and tail classes, respectively. Let the classes ordered from head to tail be $c_0,c_1,\ldots,c_{C-1}$. Given a target imbalance ratio, the target size of the class at rank $r$ is
\begin{equation}
n_r^{\mathrm{target}}
=
\left\lfloor
n_{\max}
\left(\frac{1}{\mathrm{IR}}\right)^{\frac{r}{C-1}}
\right\rfloor .
\end{equation}
We uniformly sample without replacement
\begin{equation}
n_r^{\mathrm{keep}}
=
\min\left(n_r^{\mathrm{target}},\,|\mathrm{pool}(c_r)|\right)
\end{equation}
slides from each class-specific training pool. Consequently, the realized IR can differ slightly from the nominal value because of integer rounding or a limited class pool.

\subsection{Head, Medium, and Tail Groups}

Head, Medium, and Tail groups are determined only by the class
frequencies in the final training set and are used for evaluation
and analysis, rather than model training. Under the main settings, the class assignments are shown in Table~\ref{tab:frequency-groups}.

\begin{table}[t]
\centering
\small
\caption{Class assignments to frequency groups.}
\label{tab:frequency-groups}
\begin{tabular*}{\columnwidth}{@{\extracolsep{\fill}}lccc@{}}
\toprule
Dataset & Head & Medium & Tail \\
\midrule
Camelyon+-LT (IR=30) & 0, 2 & 1 & 3 \\
PANDA-LT (IR=50) & 0, 1 & 2, 4 & 3, 5 \\
TissueNet-LT (IR=10) & 1, 0 & 2 & 3 \\
\bottomrule
\end{tabular*}
\end{table}


We adopt an adaptive grouping protocol. If at least one class contains fewer than 20 training slides, classes with more than 100 slides are assigned to Head, those with fewer than 20 to Tail, and the remainder to Medium. Otherwise, classes are sorted by decreasing training frequency and divided into three contiguous groups as evenly as possible, with any remainder assigned to Head first and then Medium. The same grouping protocol is used in the evaluation and instance-level long-tail analyses.

\subsection{Camelyon+-LT}

Camelyon+-LT contains four lymph-node metastasis categories: Negative (label 0), macro-metastasis (label 1), micro-metastasis (label 2), and isolated tumor cells (ITC; label 3). Its original distribution is naturally imbalanced. Detailed split statistics are reported in Table~\ref{tab:camelyon-stats}.

\begin{table}[t]
\centering
\small
\caption{Camelyon+-LT data statistics.}
\label{tab:camelyon-stats}
\begin{tabular*}{\columnwidth}{@{\extracolsep{\fill}}lccccc@{}}
\toprule
Split & 0 & 1 & 2 & 3 & Total \\
\midrule
Full dataset & 870 & 174 & 251 & 54 & 1,349 \\
Train (IR=30) & 844 & 148 & 225 & 28 & 1,245 \\
Validation & 10 & 10 & 10 & 10 & 40 \\
Test & 16 & 16 & 16 & 16 & 64 \\
\bottomrule
\end{tabular*}
\end{table}

\begin{table*}[t]
\centering
\small
\caption{PANDA-LT data statistics under the main IR=50 setting.}
\label{tab:panda-stats}
\begin{tabular*}{\textwidth}{@{\extracolsep{\fill}}lccccccc@{}}
\toprule
Split & 0 & 1 & 2 & 3 & 4 & 5 & Total \\
\midrule
Full dataset & 2,603 & 2,399 & 1,209 & 1,118 & 1,124 & 1,102 & 9,555 \\
Train (IR=50) & 2,303 & 1,053 & 481 & 100 & 220 & 46 & 4,203 \\
Validation & 100 & 100 & 100 & 100 & 100 & 100 & 600 \\
Test & 200 & 200 & 200 & 200 & 200 & 200 & 1,200 \\
\bottomrule
\end{tabular*}
\end{table*}

\begin{table*}[t]
\centering
\small
\caption{PANDA-LT training-set statistics at different nominal imbalance ratios.}
\label{tab:panda-ir}
\begin{tabular*}{\textwidth}{@{\extracolsep{\fill}}ccccccccc@{}}
\toprule
Nominal IR & Label 0 & Label 1 & Label 2 & Label 3 & Label 4 & Label 5 & Train total & Realized IR \\
\midrule
10 & 2,303 & 1,453 & 909 & 365 & 578 & 230 & 5,838 & 10.0 \\
20 & 2,303 & 1,264 & 694 & 209 & 381 & 115 & 4,966 & 20.0 \\
30 & 2,303 & 1,166 & 590 & 151 & 299 & 76 & 4,585 & 30.3 \\
40 & 2,303 & 1,101 & 526 & 120 & 251 & 57 & 4,358 & 40.4 \\
50 & 2,303 & 1,053 & 481 & 100 & 220 & 46 & 4,203 & 50.1 \\
\bottomrule
\end{tabular*}
\end{table*}

The realized training IR is $844/28\approx30.1$. No artificial downsampling is applied when constructing the training split.

\subsection{PANDA-LT}

PANDA-LT is a six-class prostate cancer grading task. Label 0 denotes benign tissue, and labels 1--5 correspond to ISUP grades 1--5.

The head-to-tail order used for exponential downsampling is $(0,1,2,4,3,5)$. The realized IR of the main training split is $2{,}303/46\approx50.1$. The main split statistics and additional imbalance settings are summarized in
Tables~\ref{tab:panda-stats} and~\ref{tab:panda-ir}.

The validation and test sets are identical across these IR settings.

\subsection{TissueNet-LT}

TissueNet-LT contains four cervical pathology categories: normal/subnormal epithelium (label 0), low-grade squamous intraepithelial lesion (LSIL; label 1), high-grade squamous intraepithelial lesion (HSIL; label 2), and invasive squamous cell carcinoma (SCC; label 3). Detailed split statistics are provided in Table~\ref{tab:tissuenet-stats}.
\begin{table}[t]
\centering
\small
\caption{TissueNet-LT data statistics under the main IR=10 setting.}
\label{tab:tissuenet-stats}
\begin{tabular*}{\columnwidth}{@{\extracolsep{\fill}}lccccc@{}}
\toprule
Split & 0 & 1 & 2 & 3 & Total \\
\midrule
Full dataset & 268 & 288 & 238 & 219 & 1,013 \\
Train (IR=10) & 110 & 238 & 51 & 23 & 422 \\
Validation & 20 & 20 & 20 & 20 & 80 \\
Test & 30 & 30 & 30 & 30 & 120 \\
\bottomrule
\end{tabular*}
\end{table}

The head-to-tail order is $(1,0,2,3)$. The tail-class target is $\lfloor238/10\rfloor=23$, giving a realized IR of $238/23\approx10.3$.

\subsection{Morphology and Normal Prototypes}

For each dataset, global K-means is applied to $\ell_2$-normalized patch features from training slides to obtain $K=64$ morphology prototypes $\{H_k\}_{k=1}^{K}$. At most $K\times10{,}000$ patches are used to construct the clustering pool. Each patch is assigned to its nearest prototype according to cosine similarity.

Normal prototypes are constructed independently from label-0 training slides only. K-means with $K_n=32$ is applied to their $\ell_2$-normalized patch features. Each morphology prototype $H_k$ is matched to the cosine-nearest normal prototype $\hat q_k$. No validation or test slide is used to construct either prototype set. The numbers of label-0 training slides used to construct the normal prototypes
are summarized in Table~\ref{tab:normal-prototypes}.
\begin{table}[t]
\centering
\small
\caption{Training slides used to construct normal prototypes.}
\label{tab:normal-prototypes}
\begin{tabular*}{\columnwidth}{@{\extracolsep{\fill}}llc@{}}
\toprule
Dataset & Label-0 category & Slides \\
\midrule
Camelyon+-LT & Negative & 844 \\
PANDA-LT & Benign & 2,303 \\
TissueNet-LT & Normal/subnormal & 110 \\
\bottomrule
\end{tabular*}
\end{table}

\section{Implementation Details and Training Protocol}

\subsection{Three-Stage Training Pipeline}

DeCo-MIL contains three stages and two independently parameterized MIL models.

\begin{itemize}
\item \textbf{Judge pretraining.} A standard MIL model is trained on the original full bags for 30 epochs and then frozen as $f_{\mathrm{judge}}$.
\item \textbf{Debiased counterfactual attribution.} For every training slide, the frozen judge produces anchor assignments, counterfactual contributions, and anchor-specific masking ratios. This stage is performed without gradient computation.
\item \textbf{Final model training.} A freshly initialized $f_{\mathrm{final}}$ is trained for 30 epochs with online redundancy masking, anchor-stratified pseudo-bags, tail-aware oversampling, and consistency regularization weighted by $\beta=0.5$.
\end{itemize}


All methods use the same frozen CONCH features, data splits, and evaluation protocol. For each random seed, a separate judge model is trained once and used solely for offline counterfactual scoring. The resulting causal table is reused throughout final-model training for the same seed and is not shared across seeds. The judge is discarded after scoring, so DeCo-MIL introduces additional training-time overhead only and does not increase the parameter count or computation at inference. Training uses one WSI per batch, and final performance is reported over five independent runs.

\subsection{Frequency-Debiased Counterfactual Attribution}

Let $\pi_c=n_c/\sum_{c'}n_{c'}$ be the training-set slide frequency of class $c$. For a slide $X$, the frozen judge produces frequency-calibrated probabilities
\begin{equation}
p'(c\mid X)
=
\operatorname{softmax}\left(f_{\mathrm{judge}}(X)-\tau\log\boldsymbol{\pi}\right)_c,
\end{equation}
where $\tau=1.0$ controls the calibration strength. For each non-empty morphology anchor $k$, all its instances are replaced by the matched normal prototype $\hat q_k$, yielding $X^{\mathrm{do}(k)}$. The contribution of anchor $k$ to an arbitrary class $c$ is
\begin{equation}
{E'}_k^{\,c}
=
\log p'(c\mid X)
-
\log p'\left(c\mid X^{\mathrm{do}(k)}\right).
\end{equation}
Because the contribution is computed from frequency-calibrated log-posteriors rather than raw logits, it reflects not only the change in the target-class logit but also the intervention-induced change in the prior-adjusted normalization over all classes. Frequency calibration therefore affects both the factual and intervened predictions and enters ${E'}_k^{\,c}$ through the difference between their normalization terms, as derived below.

We select the strongest competing class from the calibrated prediction of the original slide as
\begin{equation}
\tilde y=\arg\max_{c\ne y}p'(c\mid X),
\end{equation}
and keep it fixed for all anchor interventions of that slide. The ground-truth and competing-class contributions are combined as
\begin{equation}
s_k={E'}_k^{\,y}-\lambda {E'}_k^{\,\tilde y},
\end{equation}
where $\lambda=0.6$ for all datasets. Equivalently,
\begin{equation}
s_k=\lambda\left({E'}_k^{\,y}-{E'}_k^{\,\tilde y}\right)
+(1-\lambda){E'}_k^{\,y}.
\end{equation}
The first term measures class-specific evidence relative to the strongest competing class, whereas the second preserves the absolute ground-truth contribution measured under the frequency-calibrated posterior. Thus, $\lambda$ controls the trade-off between class specificity and calibrated absolute contribution.

To make the effect of frequency calibration explicit, let $z_c(X)=f_{\mathrm{judge}}(X)_c$ and
\begin{equation}
Z_{\pi}(X)=\sum_j\exp\left(z_j(X)-\tau\log\pi_j\right).
\end{equation}
The calibrated log-probability is
\begin{equation}
\log p'(c\mid X)=z_c(X)-\tau\log\pi_c-\log Z_{\pi}(X).
\end{equation}
Writing $\Delta a=a(X)-a(X^{\mathrm{do}(k)})$, the class prior $\pi_c$ is unchanged by the intervention, and hence
\begin{equation}
{E'}_k^{\,c}=\Delta z_c-\Delta\log Z_{\pi}.
\end{equation}
Substitution into the anchor score gives
\begin{equation}
s_k=\Delta z_y-\lambda\Delta z_{\tilde y}
-(1-\lambda)\Delta\log Z_{\pi}.
\end{equation}
Therefore, with $\lambda=0.6$, the prior-dependent normalization term is retained with coefficient $1-\lambda=0.4$, confirming that frequency calibration remains active in the final anchor score.

For the set $\mathcal K$ of non-empty anchors in a slide, we use the numerically stable slide-wise normalization
\begin{equation}
\operatorname{Norm}(s_k)
=
\frac{s_k-\min_{j\in\mathcal K}s_j}
{\max_{j\in\mathcal K}s_j-\min_{j\in\mathcal K}s_j+\epsilon},
\qquad \epsilon=10^{-8}.
\end{equation}
If all anchor scores are identical, the normalized scores are zero. The normalized scores are mapped to
\begin{equation}
r_k=r_{\max}\left[1-\operatorname{Norm}(s_k)\right].
\end{equation}
Higher-contribution anchors therefore receive lower masking ratios. For normal slides, $r_k=0$. We set $r_{\max}=0.9$ for Camelyon+-LT and $r_{\max}=0.5$ for PANDA-LT and TissueNet-LT, as supported by the sensitivity analysis in Sec.~\ref{sec:hyperparameter-sensitivity}.

\section{Precise Definitions of Ablation Strategies}

\subsection{Masking Strategies}

Within each random seed, all masking strategies use the same CONCH features,
prototype assignments, frozen judge, maximum masking ratio $r_{\max}$,
online masking operator, final-model architecture, optimization settings,
and tail-aware training protocol.

\subsubsection{Counterfactual-Guided Masking}

This is the proposed strategy. The anchor ratios are obtained from the frequency-debiased counterfactual scores:
\begin{equation}
r_k^{\mathrm{causal}}=r_{\max}\left[1-\operatorname{Norm}(s_k)\right].
\end{equation}

\subsubsection{Random Masking}

Within each slide, the causal ratio vector is independently permuted:
\begin{equation}
r_k^{\mathrm{random}}=r_{\pi(k)}^{\mathrm{causal}},
\end{equation}
where $\pi$ is a uniformly sampled permutation. Random masking therefore preserves the multiset, mean, and sum of anchor-level ratios, but it does not necessarily preserve the realized patch-level masked fraction $\sum_km_kr_k/\sum_km_k$.

\subsubsection{Anchor-Size Masking}

Let $m_k$ be the number of instances assigned to anchor $k$. For non-degenerate anchor sizes,
\begin{equation}
r_k^{\mathrm{size}}
=
r_{\max}
\frac{m_k-m_{\min}}{m_{\max}-m_{\min}}.
\end{equation}
Thus, larger tissue regions receive higher masking ratios. If all non-empty anchors have the same size, all ratios are set to zero.

\subsubsection{Attention-Based Masking}

Let $a_n$ be the patch attention produced by the frozen judge and $\bar a_k=m_k^{-1}\sum_{n:z_n=k}a_n$. For non-degenerate anchor-level attention,
\begin{equation}
r_k^{\mathrm{attn}}
=
r_{\max}\left(
1-
\frac{\bar a_k-a_{\min}}{a_{\max}-a_{\min}}
\right).
\end{equation}
Anchors with lower mean attention receive higher masking ratios. If all non-empty anchors have identical mean attention, all ratios are set to zero.

\subsection{Counterfactual Constructions}

For a target anchor $k$, all assigned instances are replaced while the other instances remain unchanged:
\begin{equation}
X_n^{\mathrm{do}(k)}=
\begin{cases}
b_k,&z_n=k,\\
X_n,&\text{otherwise}.
\end{cases}
\end{equation}
The three compared constructions are summarized in Table~\ref{tab:counterfactual-constructions}.

\begin{table}[t]
\centering
\small
\caption{Definitions of the compared counterfactual constructions.}
\label{tab:counterfactual-constructions}
\begin{tabular*}{\columnwidth}{@{\extracolsep{\fill}}lc@{}}
\toprule
Construction & Replacement $b_k$ \\
\midrule
Zero replacement & $\mathbf{0}\in\mathbb{R}^{512}$ \\
Morphology-prototype replacement & $H_k$ \\
Matched-normal replacement (ours) & $\hat q_k$ \\
\bottomrule
\end{tabular*}
\end{table}
\FloatBarrier

All other Stage-1 and Stage-2 settings remain unchanged.

\subsection{Pseudo-Bag Sampling Strategies}

All pseudo-bag strategies use the same redundancy-reduced bag $X^\star$, sampling ratio $\rho=0.5$, number of pseudo-bags $M_{\mathrm{pb}}=3$, source-slide label, and training objective. They use the same sampling ratio but do not necessarily produce exactly identical pseudo-bag sizes because of per-anchor integer rounding and lower bounds.

\begin{itemize}
\item \textbf{Random sampling:} uniformly sample $\operatorname{round}(\rho|X^\star|)$ instances from the entire reduced bag.
\item \textbf{Equal-per-anchor sampling:} allocate an approximately equal quota to each non-empty anchor, independent of its original size.
\item \textbf{Anchor-stratified sampling (ours):} independently sample a fraction $\rho$ from every retained anchor while retaining at least one instance from each retained anchor.
\end{itemize}

\section{Additional Analysis}

\subsection{Hyperparameter Sensitivity}
\label{sec:hyperparameter-sensitivity}

\begin{figure*}[t]
  \centering
  \includegraphics[width=\textwidth]
  {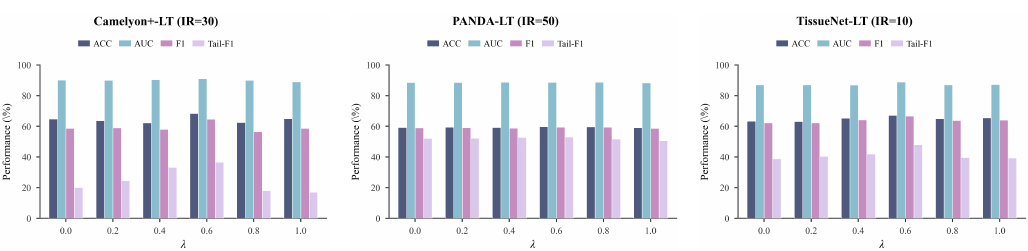}
  \caption{Sensitivity to $\lambda$. Each bar shows the mean over five runs.}
  \label{fig:lambda-sensitivity}
\end{figure*}

\begin{figure*}[t]
  \centering
  \includegraphics[width=\textwidth]
  {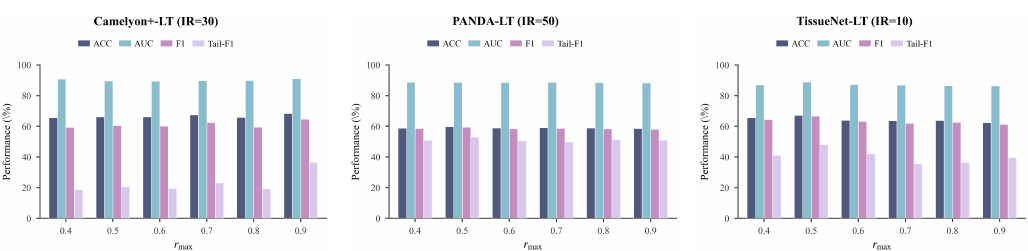}
  \caption{Sensitivity to the maximum masking ratio $r_{\max}$. Each bar shows the mean over five runs.}
  \label{fig:rmax-sensitivity}
\end{figure*}

\begin{figure*}[t]
  \centering
  \includegraphics[width=\textwidth]
  {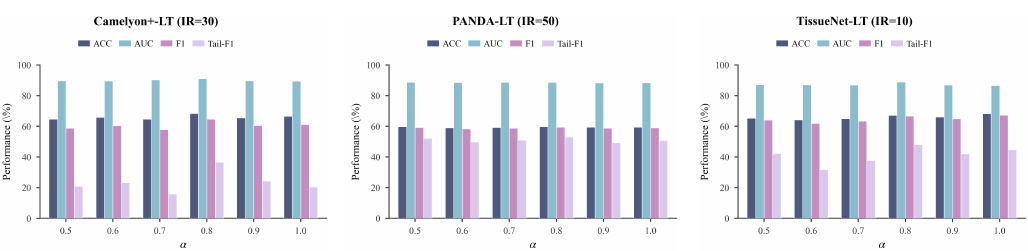}
  \caption{Sensitivity to the oversampling strength $\alpha$. Each bar shows the mean over five runs.}
  \label{fig:alpha-sensitivity}
\end{figure*}

\subsubsection{Sensitivity to $\lambda$}

We evaluate six settings on all three datasets: $\lambda=0$, $0.2$, $0.4$, $0.6$, $0.8$, and $1.0$. Each bar is the mean over five runs, and the main experiments use the unified setting $\lambda=0.6$. The corresponding results are shown in Fig.~\ref{fig:lambda-sensitivity}.

The unified setting $\lambda=0.6$ achieves the highest macro-F1 and Tail-F1 on all three datasets. In particular, $\lambda=1$ retains only the class-specific contrast and removes the calibrated absolute-contribution term in the equivalent decomposition, whereas $\lambda=0.6$ balances both components. The result supports retaining a nonzero absolute contribution without discarding competing-class suppression.


To isolate the effect of frequency calibration, ``+ Counterfactual'' in Table~2 of the main paper uses $\tau=0$, whereas ``+ Debiasing'' uses $\tau=1$. Both settings fix $\lambda=0.6$ and keep all other components unchanged. The results show that frequency calibration improves Tail-F1 on Camelyon+-LT and TissueNet-LT, while improving the overall metrics on PANDA-LT with its Tail-F1 largely preserved. These results support the effectiveness of frequency calibration in mitigating class-frequency bias.

\subsubsection{Sensitivity to $r_{\max}$}

We sweep six settings on all three datasets: $r_{\max}=0.4$, $0.5$, $0.6$, $0.7$, $0.8$, and $0.9$. Each bar is the mean over five runs. The corresponding results are shown in Fig.~\ref{fig:rmax-sensitivity}.

Camelyon+-LT benefits from stronger masking and reaches its best macro-F1 and Tail-F1 at $r_{\max}=0.9$. For PANDA-LT and TissueNet-LT, $r_{\max}=0.5$ provides the best balance and is used in the main experiments.  The maximum masking ratio $r_{\max}$ is therefore selected at the
dataset level based on the sensitivity analysis and then fixed across
all runs and comparisons. Unless otherwise specified, all other
hyperparameters are shared across datasets. The different optima reflect dataset-dependent redundancy and should not be interpreted as a universal masking strength.

\subsubsection{Sensitivity to $\alpha$}

We sweep six oversampling settings on all three datasets: $\alpha=0.5$, $0.6$, $0.7$, $0.8$, $0.9$, and $1.0$. Each bar is the mean over five runs. The corresponding results are shown in Fig.~\ref{fig:alpha-sensitivity}.


We use the unified setting $\alpha=0.8$, which gives the strongest overall balance on Camelyon+-LT and PANDA-LT and the highest Tail-F1 on TissueNet-LT. Although $\alpha=1.0$ yields a slightly higher macro-F1 on TissueNet-LT, its Tail-F1 is lower than that obtained with $\alpha=0.8$. Overall, performance remains relatively stable across a broad range of $\alpha$, with $\alpha=0.8$ providing the most favorable balance between overall metrics and Tail-F1.

\subsection{Nested Dual Long-Tail Statistics}

For PANDA-LT at IR=50, let $\mathcal S_t$ denote the pooled lesion slides whose classes belong to frequency group $t\in\{\mathrm{Head},\mathrm{Medium},\mathrm{Tail}\}$. For anchor $k$, let $\mathcal S_{t,k}\subseteq\mathcal S_t$ contain only slides in which that anchor is present and has a valid contribution. The group-level contribution is the slide-wise mean
\begin{equation}
\bar E_k^t
=
\frac{1}{|\mathcal S_{t,k}|}
\sum_{i\in\mathcal S_{t,k}}{E'}_{ik}^{\,y_i}.
\end{equation}
Thus, slides are equally weighted within each group. An absent anchor is stored as NaN and does not enter the denominator for that anchor. If $\mathcal S_{t,k}$ is empty, anchor $k$ is omitted from the group-level statistics. Normal slides are also omitted because no counterfactual contribution is stored for them. The ranked contribution curve uses the raw signed values of $\bar E_k^t$.

For the concentration statistics, we define $u_k^t=\max(\bar E_k^t,0)$ over all valid group-level anchors. The Gini coefficient for the resulting non-negative vector $u=(u_1,\ldots,u_n)$, sorted in ascending order, is
\begin{equation}
G(u)=\frac{2\sum_{i=1}^{n}i u_i}{n\sum_{i=1}^{n}u_i}-\frac{n+1}{n}.
\end{equation}
The top-10\% set contains $\max(1,\operatorname{round}(0.1n))$ anchors with the largest raw $\bar E_k^t$. Its contribution share is the sum of their non-negative contributions divided by $\sum_k u_k^t$. Patch coverage is the number of patches assigned to these top-ranked anchors, pooled over the same lesion slides, divided by the total number of patches assigned to all valid anchors. In the observed results, the top 10\% of anchors account for 53\%--74\% of the total positive contribution while covering only 15\%--22\% of patches, and the Gini coefficient increases from Head to Tail.

The corresponding ranked-contribution and concentration curves are shown in Figure~5 of the main paper.

\subsection{Additional Visualization Results}

\begin{figure*}[t]
    \centering
    \includegraphics[width=\textwidth]    {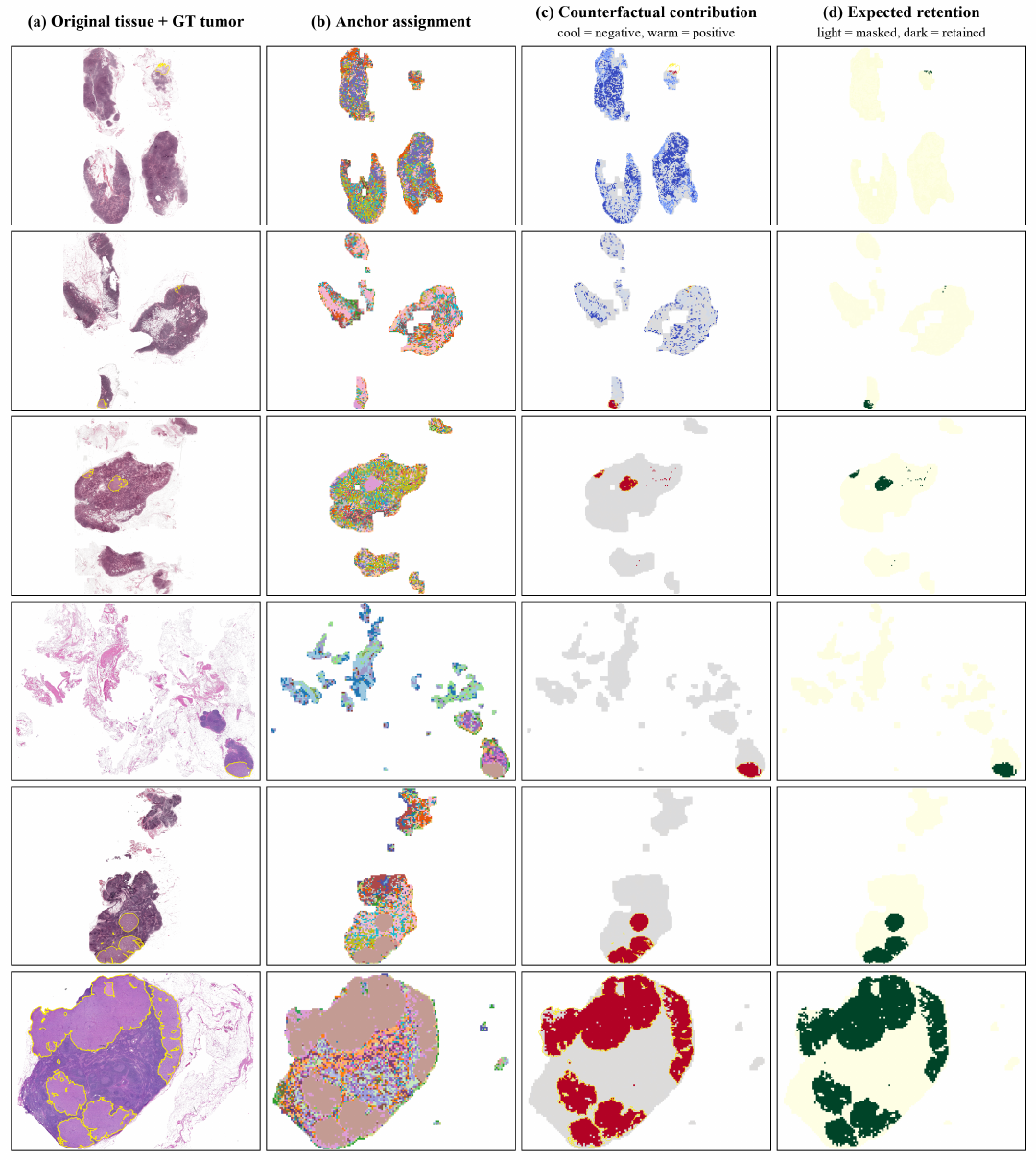}
    \caption{Additional visualizations on Camelyon+-LT. Each row
    shows a different correctly classified slide. From left to right:
    the original tissue image, morphology-anchor assignment, debiased
    counterfactual contribution, and expected retention map. Yellow
    contours denote official tumor annotations. Warm and cool colors
    indicate positive and negative contributions, respectively, while
    darker green indicates higher expected retention.}
    \label{fig:additional-visualization}
\end{figure*}

Figure~\ref{fig:additional-visualization} extends the interpretability
analysis in Fig.~6 of the main paper with additional correctly
classified slides from Camelyon+-LT, covering small, fragmented, and
extensive lesions. Despite substantial variations in tissue morphology
and lesion extent, regions with high positive counterfactual
contributions generally align with the official tumor annotations,
whereas most non-lesion regions exhibit near-zero or negative
contributions. These high-contribution regions are consequently assigned
higher expected retention, while less informative tissue is more likely
to be removed during redundancy masking. A consistent correspondence among tumor annotations, counterfactual
contributions, and retention maps is also observed across these
additional cases, further complementing the visualization results
presented in the main paper. Overall, these
results further illustrate how DeCo-MIL identifies lesion-related tissue
morphologies and preserves diagnostically informative evidence under
long-tailed WSI classification.

\end{document}